\documentclass[letterpaper]{article}
\usepackage{aaai24}
\usepackage{times}
\usepackage{helvet}
\usepackage{courier}
\usepackage[hyphens]{url}
\usepackage{graphicx}
\usepackage{natbib}
\usepackage{caption}
\usepackage{algorithm}
\usepackage{algorithmic}
\usepackage{newfloat}
\usepackage{listings}

\usepackage{url}
\usepackage{booktabs}
\usepackage{amsfonts}
\usepackage{nicefrac}
\usepackage{microtype}
\usepackage{xcolor}

\graphicspath{ {./images/} }
\usepackage{comment}

\usepackage{tikz}
\usepackage{pgfplots}
\usepackage{pgfplotstable}
\usetikzlibrary{patterns}
\usepgfplotslibrary{colormaps}
\pgfplotsset{compat=1.18}
\usetikzlibrary{pgfplots.colormaps}
\usepackage{pgf-pie} 
\usepackage{xcolor,colortbl}
\usepackage{multirow}
\usepackage{xurl}
\usepackage{siunitx}

\usepackage{dcolumn}
\newcolumntype{d}[1]{D{.}{.}{#1}}

\usepackage{makecell}
\newcommand*{\img}[1]{%
    \raisebox{-.3\baselineskip}{%
        \includegraphics[
        height=\baselineskip,
        width=\baselineskip,
        keepaspectratio,
        ]{#1}%
    }%
}

\DeclareCaptionStyle{ruled}{labelfont=normalfont,labelsep=colon,strut=off} 
\floatstyle{ruled}
\newfloat{listing}{tb}{lst}{}
\floatname{listing}{Listing}

\title{ML-EAT: A Multilevel Embedding Association Test  for Interpretable and Transparent Social Science}

\author {
    Robert Wolfe,
    Alexis Hiniker,
    Bill Howe
}
\affiliations {
    University of Washington\\
    rwolfe3@uw.edu, alexisr@uw.edu, billhowe@uw.edu
}

\begin{document}

\maketitle

\begin{abstract}
This research introduces the Multilevel Embedding Association Test (ML-EAT), a method designed for interpretable and transparent measurement of intrinsic bias in language technologies. The ML-EAT addresses issues of ambiguity and difficulty in interpreting the traditional EAT measurement by quantifying bias at three levels of increasing granularity: the differential association between two target concepts with two attribute concepts; the individual effect size of each target concept with two attribute concepts; and the association between each individual target concept and each individual attribute concept. Using the ML-EAT, this research defines a taxonomy of EAT patterns describing the nine possible outcomes of an embedding association test, each of which is associated with a unique EAT-Map, a novel four-quadrant visualization for interpreting the ML-EAT. Empirical analysis of static and diachronic word embeddings, GPT-2 language models, and a CLIP language-and-image model shows that EAT patterns add otherwise unobservable information about the component biases that make up an EAT; reveal the effects of prompting in zero-shot models; and can also identify situations when cosine similarity is an ineffective metric, rendering an EAT unreliable. Our work contributes a method for rendering bias more observable and interpretable, improving the transparency of computational investigations into human minds and societies.
\end{abstract}
\section{Introduction}

Computational methods that quantify societal biases using language technologies like word embeddings \cite{mikolov2013distributed}, generative language models \cite{radford2018improving}, and multimodal language-and-image models \cite{radford2021learning} have been widely adopted by social scientists, who leverage the reflection of society captured by these technologies to observe implicit and explicit societal biases where human-subjects experiments are infeasible or prohibitively expensive \cite{durrheim2023using, kennedy2021text}. Social scientists have employed word embeddings in particular to analyze diachronic historical changes in human biases and norms~\cite{garg2018word}, to study variations in gender biases across numerous languages \cite{lewis2020gender}, to compare implicit biases in adult and children's language corpora \cite{charlesworth2021gender}, and to validate longstanding theories about societal biases, such as the masculine default \cite{caliskan2022gender, bailey2022based}. \citet{bhatia2023predicting} employ such techniques to \textit{predict} human biases, potentially facilitating mitigations at the societal scale. Among the most widely adopted bias measurement methods in computational social science is the Word Embedding Association Test (WEAT) \cite{caliskan2017semantics}, a statistical technique grounded in the Implicit Association Test (IAT), a widely used measurement of unconscious bias in human subjects \cite{greenwald1998measuring}. The WEAT quantifies bias based on the differential cosine similarity of two target groups $X$ and $Y$ (such as Science vs. Art) with two attribute groups $A$ and $B$ (such as Male vs. Female, for a common test of gender bias).

\begin{figure*}
\centering
\includegraphics[width=.81\textwidth]{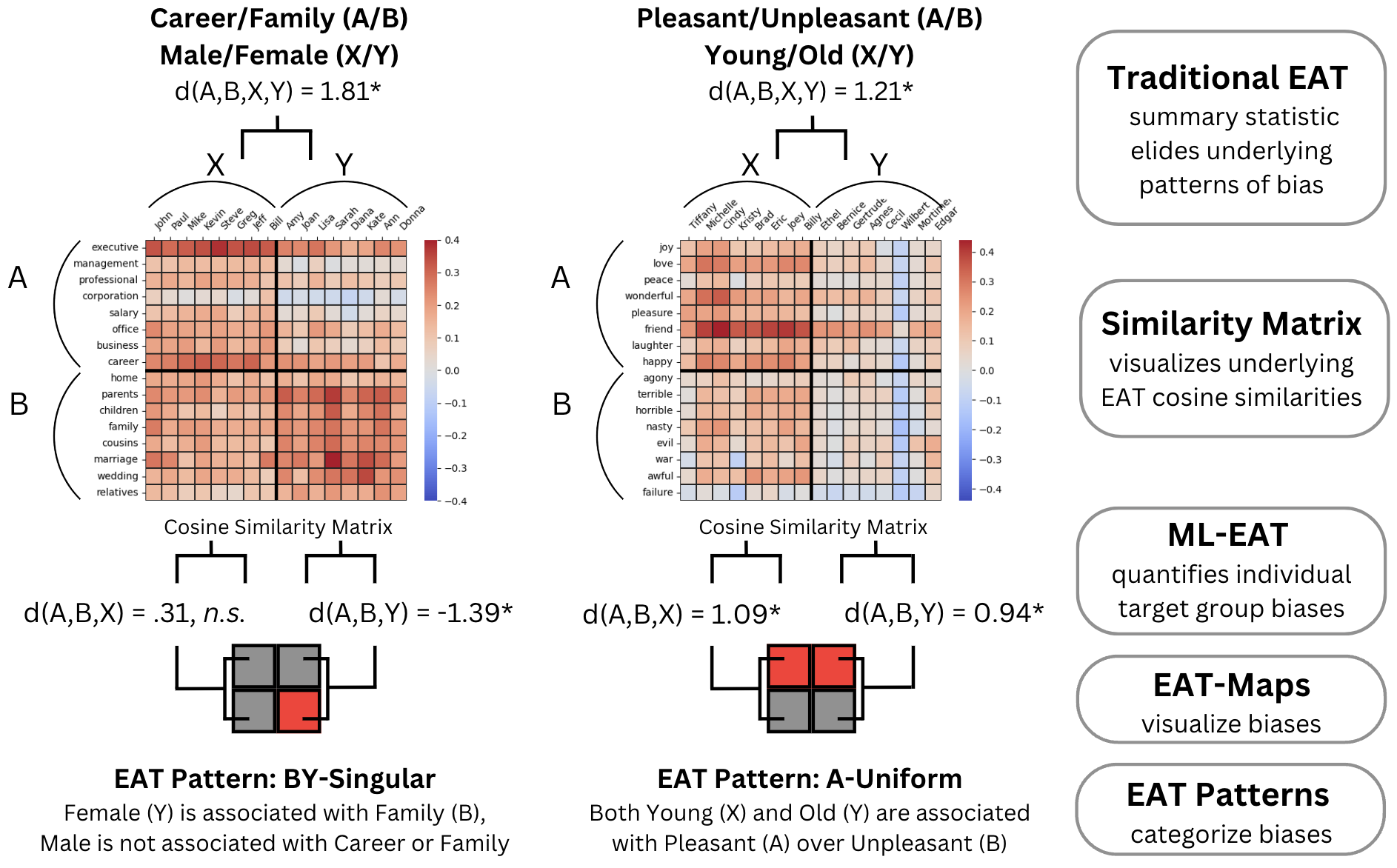}
\caption{\footnotesize A visualization of the ML-EAT applied to the Career/Family and Young/Old EATs introduced by \citet{caliskan2017semantics}. Where traditional EATs return a single effect size and $p$-value, the ML-EAT surfaces underlying patterns of bias in individual target group associations with the A or B attribute. A taxonomy of EAT patterns and EAT-Map visualizations provide a categorical and visual vocabulary for describing differences between EATs. \looseness=-1}
\label{fig:mleat_process}
\end{figure*}

Yet the WEAT also suffers from a limitation common in AI evaluation: it is an aggregate metric \cite{burnell2023rethink}, averaging over many sub-measurements between groups of words to produce a single summary statistic. Explaining a bias quantified by the WEAT is thus not straightforward. In plain English, a statistically significant WEAT indicates that the $X$ target group is more associated with $A$ relative to $B$ than the $Y$ target group is more associated with $A$ relative to $B$. While effective for surfacing implicit biases, the method also raises questions about whether and to what extent each target group is \textit{individually} associated with $A$ or $B$. Consider the test of age bias presented by \citet{caliskan2017semantics}, which sets group $A$ to Pleasantness, $B$ to Unpleasantness, $X$ to Young Names, and $Y$ to Old Names and returns a large, statistically significant effect size of 1.21. While one might intuitively interpret the result of the test to mean that Young Names are associated with Pleasantness, and Old Names with Unpleasantness, examining component cosine similarities used to compute the WEAT reveals that \textit{both} Young Names ($X$) and Old Names ($Y$) are associated with Pleasantness ($A$). On the other hand, consider the WEAT setting $A$ to Pleasantness, $B$ to Unpleasantness, $X$ to Instruments, and $Y$ to Weapons, which also returns a large, significant effect size, of 1.53. Inspection of component cosine similarities reveals that, in this case, Instruments are associated with Pleasantness, while Weapons are associated with Unpleasantness. Though the two WEATs return effect sizes with the same sign and similarly large magnitudes, the characteristics of their underlying biases, and the corresponding sociological interpretations of the results, differ significantly. Given the wide range of psychological and computational studies in which WEAT scores are utilized, knowing the difference between these outcomes can be important - but they are indistinguishable given a single WEAT $d$-score.

The present work provides a method and a vocabulary for describing meaningful differences in the underlying biases quantified by EATs (Embedding Association Tests, which are now widely utilized beyond word embeddings \cite{steed2021image}). We make the following contributions:

\begin{enumerate}
\item The \textbf{MultiLevel Embedding Association Test (ML-EAT)}, a formal, theoretically justified measure to disambiguate potential sources of bias in machine learned representations. The ML-EAT quantifies bias at three levels: Level 1, an effect size describing the differential association between two target concepts with two attribute concepts (equivalent to a traditional WEAT); Level 2, two effect sizes describing the differential association of each target group $X$ and $Y$ \textit{individually} with the two attribute groups $A$ and $B$; and Level 3, the means and standard deviations of the four underlying distributions of cosine similarities $(X,A)$, $(X,B)$, $(Y,A)$, $(Y,B)$ that ultimately compose an EAT measurement.

\item \textbf{A taxonomy of EAT patterns} for characterizing the nine discrete outcomes of EATs, each of which admits a distinct interpretation. The taxonomy provides a vocabulary for the underlying associations that make up an EAT, based on whether each of the target groups $X$ and $Y$ individually exhibits a significant bias toward $A$ or $B$.

\item \textbf{The EAT-Map, an intuitive visualization for interpreting the ML-EAT}. As illustrated near the bottom of Figure \ref{fig:mleat_process}, the EAT-Map uses a four-quadrant square with columns corresponding to the target groups $X$ and $Y$ and rows corresponding to the attribute groups $A$ and $B$, and shades cells in order to denote associations between a target group and an attribute. The EAT-Map provides a visual vocabulary for the taxonomy of EAT patterns, as each EAT pattern has a unique EAT-Map.

\item \textbf{An empirical analysis of the ML-EAT} applied to static and diachronic word embeddings \cite{pennington2014glove,hamilton2016diachronic}, GPT-2 generative language models \cite{radford2019language}, and a CLIP language-and-image model \cite{radford2021learning}. Applying the ML-EAT to the GloVe embeddings shows that five distinct EAT patterns occur in the ten WEATs performed by \citet{caliskan2017semantics}, while using the test with the HistWords embeddings \cite{hamilton2016diachronic} shows that EAT patterns help to draw more complete conclusions about historical biases. Analysis of GPT-2 and CLIP models further demonstrates that the ML-EAT can surface the effects of prompting in the zero-shot setting, as well as identify embedding spaces unsuitable for analysis with cosine similarity. \looseness=-1

\end{enumerate}

\noindent The ML-EAT provides an expressive means to describe bias in language technologies. As we will demonstrate, even well-studied results, such as biases measured by \citet{caliskan2017semantics}, yield new perspectives with the ML-EAT. Our research code is available at https://github.com/wolferobert3/ml-eat.
\section{Related Work}
In reviewing the related on work on Embedding Association Tests (EATs), we describe the models and domains in which EATs are employed; then provide a detailed overview of their applications in computational social science; and finally review their limitations as described in prior work.

\subsection{Embedding Association Tests}

\citet{caliskan2017semantics} introduce the Word Embedding Association Test (WEAT), a measurement of intrinsic bias in word embeddings drawing on the design of the Implicit Association Test (IAT), a method for studying implicit (unconscious) bias in human subjects \cite{greenwald1998measuring}. The WEAT quantifies the relative association of two target groups (such as Science and Art) with two attribute groups (such as Male and Female), and, like the IAT, returns an effect size (Cohen's $d$) and a $p$-value. \citet{caliskan2017semantics} used the WEAT to replicate the results of ten IATs reflecting gender and racial biases, among others, in the GloVe embeddings of \citet{pennington2014glove}. \citet{caliskan2017semantics} also introduced the Single Category SC-WEAT, which quantifies the differential association of a single word with two attribute groups. The SC-WEAT was introduced as part of a measurement called the Word Embedding Factual Association Test, and further clarified in an analysis of androcentric gender bias by \citet{caliskan2022gender}.

The WEAT was extended to transformer models by \citet{may2019measuring}, who introduced the Sentence Embedding Association Test (SEAT), a sentence-level WEAT employing semantically neutral prompts, and by \citet{kurita2019measuring}, who tied bias measurement to the model's pretraining objective. \citet{guo2021detecting} introduced the Contextualized Embedding Association Test (CEAT), which measured embedding bias at the word level and modeled contextualization as a random effect. EATs are also used in computer vision and multimodal language-and-image models. \citet{steed2021image} introduced the Image Embedding Association Test (iEAT), which quantified bias in self-supervised image encoders such as SimCLR \cite{chen2020simple} and iGPT \cite{chen2020generative}. \citet{ross2020measuring} introduced the Grounded-WEAT for grounded language-and-image models, while \cite{wolfe2022american} use an EAT to quantify bias in CLIP language-and-image models \cite{radford2021learning}, and \citet{hausladen2024} employ the SC-EAT to perform a causal analysis of social bias in language-and-image models. Finally, \citet{slaughter2023pre} introduce the SpEAT, an EAT for quantifying intrinsic bias in pretained speech processing models such as wav2vec2 \cite{baevski2020wav2vec} and OpenAI Whisper \cite{radford2023robust}. \looseness=-1

While most EATs employ cosine similarity to measure association between target and attribute stimuli, recent work explores alternatives. \citet{omrani2023evaluating} employ an algebraic definition of bias similar to that of \citet{bolukbasi2016man}, but use an EAT-based formula to obtain an effect size and $p$-value. \citet{bai2024measuring} assess implicit bias using the textual output of ostensibly debiased generative language models by employing a prompt-based analogue of the IAT.

\subsection{Applications of EATs in Social Science}\label{eatapplications}

Social scientists use word embeddings and EATs to study phenomena that are impossible or financially infeasible to measure solely through direct experimental methods. Many studies leverage the societal scale of the data used for training word embeddings to make broad inferences about society that would be unavailable in a small-$N$ psychological study. \citet{caliskan2022gender} demonstrate a masculine default in the English-language internet using EATs computed for every word in the vocabulary of pretrained GloVe and FastText embeddings, while \citet{bailey2022based} use word embeddings to demonstrate the implicit equivalence of the concept of ``person'' with ``men.'' \citet{napp2023gender} use EAT measurements to contend that gender stereotypes are stronger in countries that are more economically developed and individualistic. Finally, \citet{schmahl-etal-2020-wikipedia} use the EAT to study changes in gender bias on Wikipedia, informing their suggestions for reducing bias on the platform. Other research employs word embeddings for previously untenable cross-cultural analyses of human attitudes. For example, \citet{mukherjee2023global} employ EAT to measure biases related to albeism, immigration, and education across 24 languages.

Embeddings of historical data have also provided a means to study human societies that no longer exist and are thus unavailable for direct study. \citet{charlesworth2022historical} use word embeddings to measure changes in stereotypes about social groups over 200 years. \citet{borenstein2023measuring} use the EAT to study intersectional biases in word embeddings trained on newspapers from the 18th and 19th century. \citet{sunsay2023historical} use the EAT to study the disease avoidance theory of xenophobia, measuring EAT scores in 19th and 20th century travel literature to test western associations of indigenous people disgust words that would suggest disease avoidance. \citet{leach2023word} employ the EAT to argue that ``language has changed in a way that reflects greater concern for others,'' supporting the idea that the societal ``moral circle'' expanded during the 19th and 20th centuries to include more groups of people, in addition animals and the environment.  \citet{guan2024have} use cosine similarities to measure the evolution of color associations (\textit{e.g.,} association of the color red with heat) over 200 years, while \citet{betti2023large} use the EAT to measure sexism in fifty years of English song lyrics. Finally, \citet{wolfe2022hypodescent} find evidence of the historical bias of hypodescent in CLIP models.

Research also employs EATs to study present-day bias in domains such as law and medicine. \citet{rios2020quantifying} use the WEAT to measure gender bias in biomedical research, finding that traditional gender stereotypes have declined over time, but that specific medical conditions like body dysmorphia still exhibit high gender bias. \citet{cobert2024measuring} use cosine similarities to measure implicit racial biases in ICU notes. In the legal domain, \citet{matthews2022gender} use the EAT to study biases in word embeddings trained on corpora of legal opinions, and \citet{dutta2023disentangling} use WEAT scores to quantify gender bias in Indian divorce court proceedings. Moreover, amid increasing interest in using AI to measure aspects of human society \cite{park2023generative,shanahan2023role,xu2023exploring},  scholars have used EATs in an attempt to \textit{predict} human attitudes. For example, \citet{bhatia2023predicting} use WEAT scores and a novel Valence Estimation Model (VEM) to predict human implicit biases. Similarly, \citet{morehouse2023traces} measure the relationship of implicit and explicit human biases using the IAT, the WEAT, and the Mean Average Cosine method \cite{manzini2019black}. \looseness=-1

\subsection{Limitations of EATs}\label{sec:eatlimitations}

The EAT faces challenges related both to its mathematical definition and its predictive value in NLP applications. Social scientists who choose not to use the EAT sometimes note that its design, which mimics the differential construction of the IAT, can cause difficulties in observing and interpreting bias. \citet{bailey2022based} use the difference in raw cosine similarities to observe asymmetrical gender biases, noting that the WEAT is better applied to symmetrical patterns of association. \citet{ethayarajh2019understanding} contend that the standardization of the WEAT, in dividing by the joint standard deviation of word associations, can obscure differences in underlying cosine similarities. Moreover, anisotropy (directional uniformity) in deep learning models \cite{mu2018all} can distort intrinsic semantic measurements \cite{timkey2021all}, necessitating postprocessing of EATs \cite{wolfe2022vast}.

Recent work suggests that EATs have limited predictive value for bias in downstream NLP tasks. \citet{goldfarb2020intrinsic} find that biases observed with the WEAT are not correlated with application biases in tasks like coreference resolution. In a study of downstream propagation using transformer language models, \citet{orgad2022gender} propose an information-theoretic framework rather than an EAT. \citet{cabello2023independence} show that association bias and fairness are uncorrelated, but also provide sociological evidence that the two kinds of metrics should be expected to be independent of each other.

We are concerned with uses of the EAT in social science to study human attitudes. To that end, we design an interpretable EAT, rather than an EAT predictive of downstream bias.
\section{Models and Data}\label{sec:data_models}

This research introduces the ML-EAT and applies it to word embeddings, GPT-2, and CLIP, adapting stimuli from prior studies of implicit bias in NLP and computer vision. \looseness=-1

\subsection{Pretrained Models}

We apply the ML-EAT to the below language technologies.

\begin{itemize}
\item \textbf{GloVe Word Embeddings}: \textbf{Glo}bal \textbf{Ve}ctors for word representation (GloVe) train on the co-occurrence matrix of a text corpus, such that the vector representation of a word is learned based on the words it is most likely to occur around \cite{pennington2014glove}. \citet{caliskan2017semantics} introduced the WEAT by presenting results on 300-dimensional GloVe vectors trained on the 840-billion token Common Crawl. \looseness=-1

\item \textbf{HistWords Embeddings}: HistWords refers to sets of 20 word embeddings in four languages trained on ten-year slices of historical language corpora ranging between the years 1800 and 2000 \cite{hamilton2016diachronic}. \citet{hamilton2016diachronic} introduced HistWords to prove that more frequently used words exhibit less semantic change over time, and that polysemous words exhibit faster semantic change. We apply the ML-EAT to the English language HistWords embeddings trained using Word2Vec (SGNS) \cite{mikolov2013distributed} on Google books (all genres) \cite{lin2012syntactic}.

\item \textbf{GPT-2 Language Models:} GPT-2 (``Generative Pretrained Transformer'') is a causally masked transformer \cite{vaswani2017attention} language model trained to predict the next word in a sequence  \cite{radford2019language}. This research studies the four pretrained GPT-2 models (Base, Medium, Large, and XL) available via the Transformers library \cite{wolf-etal-2020-transformers}, which were pretrained on OpenAI's WebText dataset, a collection of webpages scraped from highly rated outbound links on Reddit. 

\item \textbf{CLIP Language-and-Image Models:} CLIP (``Contrastive Language Image Pretraining'') is a multimodal language-and-image model, which classifies images based on their cosine similarity with text labels \cite{radford2021learning}. This research reports results under varying prompts from the CLIP-ViT-L14-336 model, the best performing OpenAI-trained CLIP model available. CLIP-ViT-L14-336 is trained on OpenAI's WebImageText (WIT) dataset, a collection of 400 million pairs of web-scraped images and accompanying captions \cite{radford2021learning}.

\end{itemize}

\noindent GPT-2 embeddings are obtained from the model's top layer, consistent with both \citet{may2019measuring} and \citet{guo2021detecting}. CLIP embeddings are collected after projection to the model's multimodal text-and-image latent space. 

\subsection{EAT Stimuli}

An EAT employs four groups of words or images (called ``stimuli'', drawing on the test's psychological foundations in the IAT \cite{greenwald1998measuring}), each representing a concept. For example, the EAT demonstrating that flowers are favored over insects uses word lists to represent the concepts of Flowers, Insects, Pleasant, and Unpleasant. Each EAT includes two ``target'' groups, $X$ and $Y$ (Flowers and Insects), which are tested for association with two ``attribute'' groups, $A$ and $B$ (Pleasant and Unpleasant) \cite{caliskan2017semantics}. The two target groups contain the same number of stimuli, as do the two attribute groups. Groups must contain at least eight stimuli to adequately represent a concept \cite{caliskan2022gender}.

We use the stimuli for the tests of implicit bias specified by \citet{caliskan2017semantics} when applying the ML-EAT to GloVe and GPT-2. We applied the Math/Arts Male/Female EAT to the HistWords embeddings, replacing three stimuli because their L2 norms were zero-valued (preventing the computation of cosine similarity) in several HistWords embeddings. We substituted ``music'' for ``symphony''; ``mathematics'' for ``math''; and ``calculation'' for ``calculus.'' Tests of bias in CLIP utilize the word stimuli of \citet{caliskan2017semantics} to represent Pleasant and Unpleasant, and image stimuli from \citet{steed2021image}.
\begin{figure*}
\centering
\includegraphics[width=.8\textwidth]{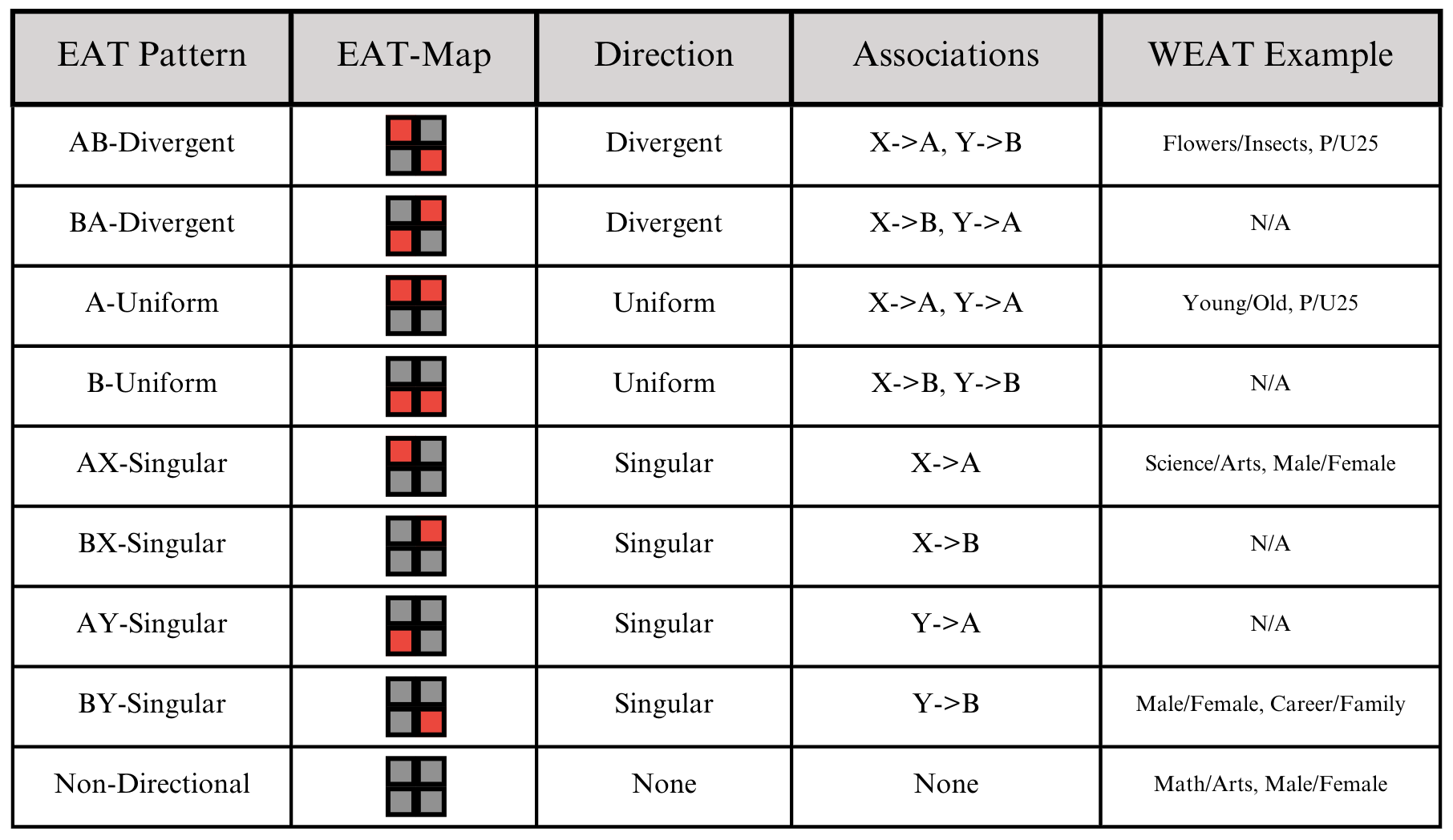}
\caption{\footnotesize A taxonomy of EAT patterns to describe the associations of an EAT's target groups with its two attribute groups. Each pattern has a unique EAT-Map formed by shading cells corresponding to significant Level 2 tests, with target groups on the X-axis and attributes on the Y. \looseness=-1}
\label{eat_patterns}
\end{figure*}

\section{Approach}\label{sec:approach}

The ML-EAT is defined using three levels of measurement, with a taxonomy of nine EAT patterns for describing biases it quantifies. We first describe the test itself, then introduce the EAT-Map visualization and EAT pattern taxonomy.

\subsection{Defining the ML-EAT}

The ML-EAT computes bias at three levels of increasing granularity. Level 1 returns the traditional standardized effect size quantifying the differential association between two target concepts with two attribute concepts; Level 2 returns two effect sizes quantifying the differential association of each target group individually with the two attribute groups; and Level 3 returns four means and corresponding standard deviations describing the non-differential association of each target group and attribute group.

\subsubsection{Level 1}

The first level of the ML-EAT is equivalent to the WEAT, as given by \citet{caliskan2017semantics}:
\begin{equation}
\frac{\textrm{mean}_{x\in X}s(x,A,B) - \textrm{mean}_{y\in Y}s(y,A,B)}{\textrm{std\_dev}_{w \in X \cup Y}s(w,A,B)}
\end{equation}
where $A$ and $B$ are attributes, $X$ and $Y$ are target groups, and association \textit{s()} for an embedding $\vec{w}$ is:
\begin{equation}
{\textrm{mean}_{a\in A}\textrm{cos}(\vec{w},\vec{a}) - \textrm{mean}_{b\in B}\textrm{cos}(\vec{w},\vec{b})}
\end{equation}

\noindent This means that the association \textit{s()} for each target stimulus represented by $\vec{w}$ is equal to its average association with the attribute stimuli in $A$, minus its average association with the attribute stimuli in $B$. The EAT returns an effect size (Cohen's $d$ \cite{cohen1992statistical}), and a $p$-value from a permutation test:  \looseness=-1

\begin{equation}
\textrm{Pr}_i[s(Xi,Yi,A,B) > s(X,Y,A,B)] 
\end{equation}
which shuffles the words of the target groups to determine the unlikeliness of the test statistic $s(X,Y,A,B)$, given by:

\begin{equation}
\sum_{x \in X} s(x,A,B) - \sum_{y \in Y} s(y,A,B)
\end{equation}

\subsubsection{Level 2}

We introduce Level 2 of the ML-EAT, which quantifies the differential association of a single Target concept $T$ with attributes $A$ and $B$:

\begin{equation}
\label{eq:lvl2}
d_{A,B,T} = \frac{\textrm{mean}_{a \in A}u(T,a) - \textrm{mean}_{b \in B}u(T,b)}{\textrm{std\_dev}_{x \in A \cup B}u(T, x)}
\end{equation}

\noindent where the association $u()$ for an attribute embedding $\vec{a}$ with the target group $T$ is given by:

\begin{equation}
\textrm{mean}_{t\in T}\textrm{cos}(\vec{t},\vec{a})
\end{equation}

\noindent Like Level 1, Level 2 returns an effect size (Cohen's $d$ \cite{cohen1992statistical}) and a $p$-value from a permutation test:

\begin{equation}
\textrm{Pr}_i[u(Ai,Bi,T) > u(A,B,T)] 
\end{equation}
which shuffles the words of the attribute groups to determine the unlikeliness of the test statistic, defined as:
\begin{equation}
\sum_{a \in A} u(T,a) - \sum_{b \in B} u(T,b)
\end{equation}

Note that Level 2 computes the differential association between a target group and two attributes, rather than an attribute and two targets. This design is intentional, as it allows the ML-EAT to answer whether a target group like Young names, for example, is associated with pleasantness or unpleasantness, without reference to another target group, such as Old names. In this way, Level 2 generalizes the Single-Category Embedding Association Test (SC-EAT), with which \citet{caliskan2017semantics} compute the differential association of a single word $w$ (such as a job title) with two attributes (such as male female words). Consider the formula for the SC-EAT:

\begin{equation}
\frac{\textrm{mean}_{a\in A}\textrm{cos}(\vec{w},\vec{a}) - \textrm{mean}_{b\in B}\textrm{cos}(\vec{w},\vec{b})}{\textrm{std\_dev}_{x \in A \cup B}\textrm{cos}(\vec{w},\vec{x})}
\end{equation}

When the number of words in a target group $T$ is equal to 1, the mean of its cosine similarities with an attribute word $x$ is necessarily equal to the only cosine similarity computed, such that $u(T,x)$ reduces to $cos(\vec{w},\vec{x})$, and the formula for Level 2 reduces to that of the SC-EAT. This suggests that Level 2 can be used in ways analogous to the SC-EAT, and may be more robustly representative of a concept because it uses a target group $T$, rather than a single word $w$.

\begin{table*}[htbp]
\centering
\small
\begin{tabular}
{|l||S[table-format=3.2]|S[table-format=3.2]|S[table-format=3.2]|r|r|r|r|}
 \hline
 \multicolumn{8}{|c|}{Multilevel Embedding Association Test: GloVe Embeddings} \\
 \hline
 Level & \multicolumn{1}{|c|}{Level 1} & \multicolumn{2}{|c|}{Level 2} & \multicolumn{4}{|c|}{Level 3} \\
\hline
EAT (Targets X/Y Attributes A/B) & \multicolumn{1}{|c|}{A,B,X,Y} & \multicolumn{1}{|c|}{A,B,X} & \multicolumn{1}{|c|}{A,B,Y} & \multicolumn{1}{|c|}{A,X} & \multicolumn{1}{|c|}{B,X} & \multicolumn{1}{|c|}{A,Y} & \multicolumn{1}{|c|}{B,Y} \\
 \hline
Flower/Insect P/U25 & 1.50* & \cellcolor{red!20}.60* & \cellcolor{blue!20}-.69* & .10 (.10) & .06 (.08) & .08 (.10) & .13 (.10) \\
Instrument/Weapon P/U25 & 1.53* & \cellcolor{red!20}1.15* & \cellcolor{blue!20}-.59* & .11 (.08) & .05 (.06) & .12 (.08) & .16 (.10) \\
EA/AA32 P/U25 & 1.40* & \cellcolor{gray!20}.46 & \cellcolor{gray!20}-.31 & .12 (.10) & .09 (.08) & -.01 (.08) & .00 (.07) \\
EA/AA16 P/U25 & 1.49* & \cellcolor{gray!20}.35 & \cellcolor{gray!20}-.39 & .11 (.10) & .08 (.08) & .00 (.08) & .01 (.08) \\
EA/AA16 P/U8 & 1.28* & \cellcolor{red!20}1.12* & \cellcolor{gray!20}.63 & .18 (.09) & .10 (.06) & .02 (.07) & .00 (.07) \\
Male/Female Career/Family & 1.81* & \cellcolor{gray!20}.31 & \cellcolor{blue!20}-1.39* & .18 (.09) & .16 (.05) & .09 (.09) & .23 (.06) \\
Math/Arts Male/Female & 1.05* & \cellcolor{gray!20}.38 & \cellcolor{gray!20}-.33 & .10 (.09) & .09 (.09) & .23 (.07) & .24 (.08) \\
Science/Arts Male/Female & 1.23* & \cellcolor{red!20}.83* & \cellcolor{gray!20}-.05 & .15 (.07) & .11 (.08) & .22 (.06) & .22 (.08) \\
Mental/Physical Temp/Perm & 1.38* & \cellcolor{gray!20}-.65 & \cellcolor{blue!20}-1.20* & .24 (.12) & .29 (.12) & .18 (.10) & .32 (.15) \\
Young/Old P/U8 & 1.21* & \cellcolor{red!20}1.09* & \cellcolor{red!20}.94* & .20 (.09) & .11 (.08) & .07 (.08) & .02 (.07) \\
\hline
\end{tabular}
\caption{\footnotesize The ML-EAT reveals five EAT patterns in the tests of \citet{caliskan2017semantics}: AB-Divergent (Flower/Insect; Instrument/Weapon); Non-Directional (first two European/African American; Math/Arts); AX-Singular (third EA/AA; Science/Arts); BY-Singular (Career/Family; Mental/Physical); and A-Uniform (Young/Old). Level 2 shading describes significance and sign: red denotes significant, positive; blue significant, negative; gray non-significant.} 
\label{MLEAT_Results_Table}
\end{table*}

\subsubsection{Level 3}

Level 3 describes the distribution of cosine similarities between a target group $T$ and an attribute $A$ in terms of the mean $\overline{TA}$ and standard deviation $\sigma_{TA}$:

\begin{equation}
\overline{TA} = \frac{1}{nm} \sum_{i}^{n} \sum_{j}^{m}  cos(\vec{A_i}, \vec{T_j})
\end{equation}

\begin{equation}
\sigma_{TA} =  \sqrt{\frac{\sum_{i}^{n} \sum_{j}^{m}  (cos(\vec{A_i}, \vec{T_j}) - \overline{TA})^{2}}{nm-1}}
\end{equation}
Level 3 surfaces the magnitude of absolute (non-standardized) differences between groups. Including a non-standardized component in the ML-EAT provides two important insights for social scientists. First, it transparently surfaces underlying cosine similarities, which may be positive (indicating similarity between the groups), negative (indicating dissimilarity between the groups), or zero. Second, it reveals when cosine similarity may not be a meaningful measurement for an embedding space, as in the case of an anisotropic (directionally uniform) embedding \cite{timkey2021all}, wherein all vectors point in the same direction, usually due to a few especially high-magnitude dimensions \cite{mu2018all}. While Level 2 permits more direct comparison between two groups of cosine similarities (\textit{e.g.,} $(X,A)$ and $(Y,A)$), it still subtracts and standardizes them, rendering unavailable any interpretation that might draw on the cosine similarities themselves. 

\subsection{EAT-Maps}

We introduce the EAT-Map, which visualizes EAT results using a four-quadrant square. Columns correspond to the EAT target groups, with $X$ corresponding to the first column and $Y$ to the second. Rows correspond to the attributes, with $A$ corresponding to the first row and $B$ to the second. Level 2 results determine the shading of the EAT-map. For example, the upper right quadrant, associated with row $A$ and column $Y$, is shaded red if $d_{A,B,Y} > 0.2$ (the minimal level defined by \citet{cohen1992statistical} for a ``small'' effect) with a p-value $> 0.05$. If the $X$ target group is differentially associated with $A$, the top left cell is shaded red, and the bottom left cell is shaded gray; conversely, if $X$ is differentially associated with $B$, the bottom left cell is shaded red, and the top left is shaded gray. If $X$ is associated with neither $A$ nor $B$, both cells of the left column (corresponding to $X$) are shaded gray. This is repeated for the right column, corresponding to the $Y$ target group. The EAT-Map intuitively visualizes a taxonomy of EAT patterns, discussed next. Figure \ref{eat_patterns} illustrates the nine possible EAT-Maps, each corresponding to a distinct EAT pattern. \looseness=-1

\subsection{EAT Patterns}

We introduce EAT patterns to provide a taxonomy for describing biases quantified by Level 2. EAT patterns define EAT measurements in terms of Direction, based on whether the target groups in an EAT are individually associated (\textit{i.e.,} exhibit a significant $p$-value and at least a small positive effect size) with the same attribute group, or with differing attribute groups. An EAT exhibits one of the following four categories of Direction:

\begin{itemize}

\item \textbf{Divergent}: $X$ and $Y$ are associated with differing attributes (\textit{e.g.,} $X$ with $A$, $Y$ with $B$).

\item \textbf{Uniform}: $X$ and $Y$ are associated with the same attribute (\textit{e.g.,} $X$ with $A$, and $Y$ with $A$).

\item \textbf{Singular}: Either $X$ or $Y$ is associated with an attribute, while the other target group is not (\textit{e.g.,} $X$ with $A$, $Y$ with neither $A$ nor $B$).

\item \textbf{Non-Directional}: neither $X$ nor $Y$ is associated with either $A$ or $B$.

\end{itemize}

\noindent EAT Patterns describe an EAT's target-attribute associations by prepending them to the Direction. An EAT with Singular Direction wherein the only significant Level 2 association occurs between Target $Y$ and Attribute $B$ exhibits a $BY$-Singular EAT pattern. An EAT with Uniform Direction and significant Level 2 associations both between $X$ and $A$ and between $Y$ and $A$ exhibits an $A$-Uniform EAT pattern. Describing an EAT that exhibits a Divergent pattern is slightly different: if significant Level 2 associations occur between $X$ and $A$ and between $Y$ and $B$, the EAT exhibits an AB-Divergent pattern; if associations occur between $X$ and $B$ and between $Y$ and $A$, the EAT exhibits a BA-Divergent pattern. \looseness=-1 

Societal biases quantified using Embedding Association Tests can be described more transparently by employing the vocabulary of EAT patterns; for example, describing the outcome of the Male/Female, Science/Arts EAT as AX-Singular communicates that Science ($X$) is differentially associated with Male ($A$), but Arts ($Y$) is associated neither with Male ($A$) nor with Female ($B$). The granular information provided by EAT patterns allows social scientists to observe more about the nature of a bias than can be interpreted via the single effect size and $p$-value returned by a traditional EAT. As demonstrated in the Results section and illustrated in Figure \ref{eat_patterns}, nearly all EAT patterns consistent with a positively signed Level 1 effect size do in fact occur in the tests performed by \citet{caliskan2017semantics}, indicating that the ML-EAT can provide additional insights about diverse forms of societal bias, even where EATs have been taken previously.
\section{Results}\label{sec:results}

We apply the ML-EAT to three language technologies: static and diachronic word embeddings, GPT-2 language models and CLIP language-and-image models. Our analysis shows that a wide variety of EAT patterns occur even when Level 1 effect sizes are uniformly large, positive, and statistically significant; that Level 2 effect sizes can inform the interpretation of historical biases; and that Levels 2 and 3 of the ML-EAT can provide insight into the effects of prompting, and can surface anisotropy that may render EATs unreliable. \looseness=-1

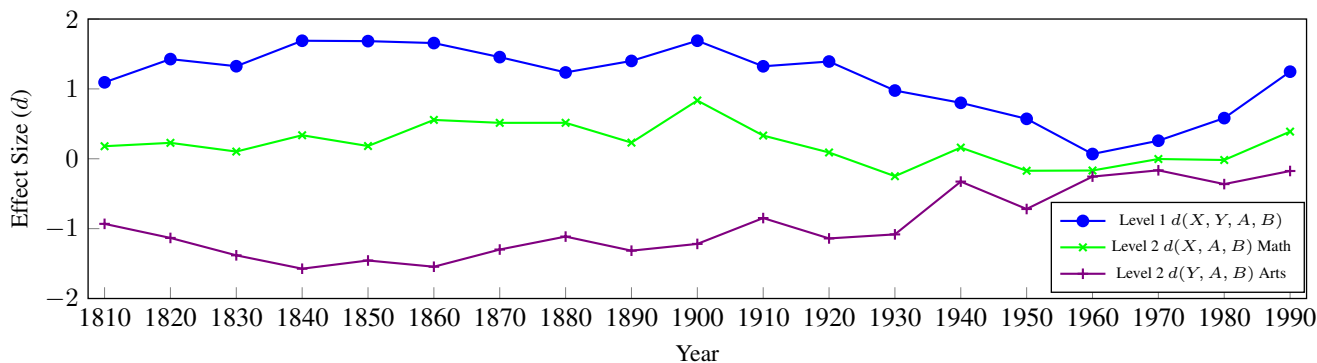
\begin{figure*}[htbp]
\begin{tikzpicture}
\begin{axis} [
    height=5.3cm,
    width=\textwidth,
    line width = .5pt,
    ymin = -2, 
    ymax = 2,
    xmin=-.25,
    xmax=18.25,
    ylabel={Effect Size ($d$)},
    ylabel shift=-2pt,
    xtick = {0,1,2,3,4,5,6,7,8,9,10,11,12,13,14,15,16,17,18},
    xticklabels = {1810,1820,1830,1840,1850,1860,1870,1880,1890,1900,1910,1920,1930,1940,1950,1960,1970,1980,1990},
    xtick pos=left,
    ytick pos = left,
    title=Math/Arts (X/Y) Male/Female (A/B) Gender Bias EAT by Decade in HistWords,
    xlabel= {Year},
    xlabel style = {font=\footnotesize},
    ylabel style = {font=\footnotesize},
    legend style={at={(.79,.02)},anchor=south west, font=\tiny}
]

\addplot[thick,solid,mark=*,color=blue] coordinates {(0, 1.0938966352531665) (1, 1.4264480834379607) (2, 1.3244630160487212) (3, 1.6891429839598318) (4, 1.6836506576751482) (5, 1.6560005920687024) (6, 1.4548428922291952) (7, 1.2357140953886758) (8, 1.4003484208211625) (9, 1.6894930161324675) (10, 1.3233783566460111) (11, 1.391919080170635) (12, 0.976555725026228) (13, 0.8007256948239234) (14, 0.5715268692252966) (15, 0.06811329954459036) (16, 0.25830912117710836) (17, 0.581878001257576) (18, 1.2465598191090386)};

\addplot[thick,solid,mark=x,color=green] coordinates {(0, 0.17961768697794056) (1, 0.22795772888319646) (2, 0.10313193716104553) (3, 0.336749603729364) (4, 0.1822002489811704) (5, 0.5560258789984741) (6, 0.5140474692362316) (7, 0.5146959892787197) (8, 0.23109552288522336) (9, 0.8335828639539974) (10, 0.33219925587798116) (11, 0.09004327864658283) (12, -0.2495682590391446) (13, 0.16002220111927623) (14, -0.17167779684391313) (15, -0.16795167074408943) (16, -0.003509783358441992) (17, -0.017649143299159866) (18, 0.3886512773623454)};

\addplot[thick,solid,mark=+,color=violet] coordinates {(0, -0.931524272669458) (1, -1.1330486435205673) (2, -1.3818656215215255) (3, -1.5735867739123532) (4, -1.4558878708702092) (5, -1.545182864001295) (6, -1.299129921169359) (7, -1.1136180929688841) (8, -1.315479451285148) (9, -1.2185009673478129) (10, -0.8485766573148521) (11, -1.1401220363533393) (12, -1.08220996780453) (13, -0.3266606834269148) (14, -0.7191082690738078) (15, -0.25525506068337067) (16, -0.1652442215145565) (17, -0.36248576972885904) (18, -0.17500033307077137)};

\legend {{Level 1 $d(X,Y,A,B)$}, {Level 2 $d(X,A,B)$ Math}, {Level 2 $d(Y,A,B)$ Arts}};
\end{axis}
\end{tikzpicture}
\caption{\small The ML-EAT can clarify underlying patterns of bias in studies of historical bias. While Math/Arts gender bias in the 1990s appears to return to 1920s magnitudes based on Level 1 effect sizes, Level 2 makes clear that the underlying bias pattern (Nondirectional, with two small, non-significant effect sizes) has not changed - although Math does exhibit a small, non-significant effect with Male in the 1990s.}
\label{fig:histwordsmatharts}
\end{figure*}

\subsection{Empirical Analysis: GloVe Embeddings}

Table \ref{MLEAT_Results_Table} presents the results of the ML-EAT applied to the ten word embedding association tests of \citet{caliskan2017semantics} and demonstrates that a wide variety of EAT patterns can result in a large, statistically significant Level 1 effect size (equivalent to the traditional WEAT effect size). Four tests exhibit Singular Direction:

\begin{itemize}
    \item The EA/AA 16 P/U 8 test exhibits an \img{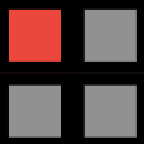} \textbf{AX-Singular} pattern, indicating that European-American is significantly associated with Pleasantness, and African-American with neither Pleasantness nor Unpleasantness.
    \item The Science/Arts Male/Female test exhibits an \img{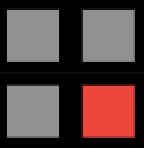} \textbf{AX-Singular} pattern, indicating that Science is associated with Male, and Arts with neither Male nor Female.
    \item The Mental/Physical Temporary/Permanent test exhibits a \img{images/by_sing} \textbf{BY-Singular} pattern, indicating that Physical is associated with Permanent, and Mental with neither Temporary nor Permanent.
    \item The Male/Female Career/Family test exhibits a \img{images/by_sing} \textbf{BY-Singular} pattern, indicating that Female is associated with Family, and Male with neither Career nor Family.
\end{itemize}

\noindent The Young/Old test exhibits an \img{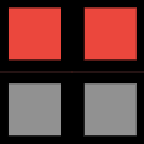} \textbf{A-Uniform pattern}, indicating that Young and Old are both differentially associated with Pleasantness; however the magnitude of association is greater for Young (1.09 vs. 0.94). The most common EAT pattern observed is \img{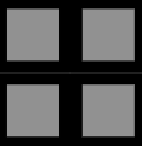} \textbf{Non-Directional}, exhibited by the first two European American/African American PU/25 tests and the Math/Arts Male/Female test. Only the Flowers/Insects P/U25 and Instruments/Weapons P/U25 tests exhibit an \img{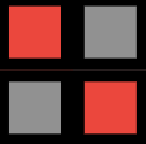} \textbf{AB-Divergent} EAT pattern, a notable finding given that discussions of EAT results often suggest this pattern (\textit{i.e.}, $X$ is differentially associated with $A$, while $Y$ is differentially with $B$). That none of the results of the \textit{social} bias tests in the GloVe embeddings exhibit an AB-Divergent EAT pattern highlights the need for descriptive reporting of EATs.

Inspection of Level 3 results reveals that small differences in cosine similarity distributions can yield large, statistically significant Level 1 effect sizes. Consider EATs exhibiting a Non-Directional pattern: in the Math/Arts, Male/Female test, the absolute difference in mean cosine similarity for Math is .01 greater with the Male attribute group (.10 vs. .09), while the absolute difference in mean cosine similarity for Arts is .01 greater with the Female attribute group (.23 vs. .24). Similarly, the mean cosine similarity for an African American names target group ($Y$ in tests 3, 4, and 5) with any attribute group never exceeds .02 or falls below -.01, suggesting a paucity of co-occurrence data for African American names due to under-representation in the training data. This is also reflected in the non-significant Level 2 effect size $A,B,Y$ for the African-American names target group. Nonetheless, Level 1 returns a large, significant effect size for these EATs.

That Level 1 picks up on small differences is a \textit{benefit} of the EAT, and Level 1 is often consistent with tests of implicit bias in humans \cite{caliskan2017semantics}. However, interpreting Level 1 without reference to Level 2 or 3 could lead to inaccurate conclusions about the direction of bias and the magnitude of absolute differences in underlying similarities between target and attribute groups.

\subsection{Empirical Analysis: HistWords Embeddings}

Among the most common uses of the EAT in computational social science is to observe change in societal biases over time \cite{charlesworth2022historical,borenstein2023measuring,betti2023large}. To illustrate how the ML-EAT can inform such studies, we quantified gender bias using the Math/Arts Male/Female EAT in the HistWords embeddings \cite{hamilton2016diachronic} from the 1810s through the 1990s (we excluded 1800 because many stimuli had zero-norm vectors, indicating insufficient co-occurrences for analysis). Figure \ref{fig:histwordsmatharts} describes Level 1 effect sizes as well as Level 2 $X,A,B$ and $Y,A,B$ effect sizes. Note that HistWords embeddings are titled with the year that starts the decade on which they train (the 1990 embedding trains on text from 1990-1999). \looseness=-1

By relying only on Level 1 (the traditional EAT), one might conclude that Science/Arts gender bias declined to its lowest level around 1960 ($d=.07$), only to increase sharply in the 1990s ($d=1.25$) to a degree not observed since the 1920s. Measuring Level 2 effect sizes adds significant context: from the 1810s through the 1890s and the 1910s through the 1930s, Arts exhibits a large, statistically significant association with Female, while Math is significantly associated with neither Male nor Female - a \img{images/by_sing} \textbf{BY-Singular} EAT pattern. The lone outlier is the 1900s, which exhibits an \img{images/ab_div} \textbf{AB-Divergent} EAT pattern (Math is significantly associated with Male, Arts with Female). Starting in the 1940s, Arts is no longer significantly associated with Female, and every EAT pattern measured thereafter corresponds to \img{images/non_dir} \textbf{Non-Directional}. While the association of Math with Male \textit{does increase} to $d=0.38$ in 1990, the effect is small and not significant; Arts remains not significantly associated with either Male or Female, with $d=-0.18$. The increased observability afforded by the ML-EAT both helps to interpret changes in bias and to prevent drawing incomplete conclusions based on the Level 1 effect size. \looseness=-1

\begin{table*}[htbp]
\centering
\small
\begin{tabular}
{|l||S[table-format=3.2]|S[table-format=3.2]|S[table-format=3.2]|r|r|r|r|}
 \hline
 \multicolumn{8}{|c|}{CLIP-ViT-L14-336 EAT Results - No Prompt} \\
 \hline
 Level & \multicolumn{1}{|c|}{Level 1} & \multicolumn{2}{|c|}{Level 2} & \multicolumn{4}{|c|}{Level 3} \\
\hline
EAT (Targets X/Y Attributes A/B) & \multicolumn{1}{|c|}{A,B,X,Y} & \multicolumn{1}{|c|}{A,B,X} & \multicolumn{1}{|c|}{A,B,Y} & \multicolumn{1}{|c|}{A,X} & \multicolumn{1}{|c|}{B,X} & \multicolumn{1}{|c|}{A,Y} & \multicolumn{1}{|c|}{B,Y} \\
 \hline
Flower/Insect P/U25 & 1.88* & \cellcolor{red!20}0.87* & \cellcolor{blue!20}-1.03* & .15 (.02) & .14 (.02) & .13 (.02) & .15 (.02) \\
White/Black P/U25 & 1.05* & \cellcolor{blue!20}-1.27* & \cellcolor{blue!20}-1.40* & .16 (.01) & .18 (.01) & .14 (.02) & .16 (.02) \\
Thin/Heavy P/U25 & 1.58* & \cellcolor{gray!20}-0.44 & \cellcolor{blue!20}-0.90* & .15 (.01) & .16 (.01) & .14 (.02) & .15 (.01) \\
Male/Female Career/Family & 0.46* & \cellcolor{red!20}1.66* & \cellcolor{red!20}1.54* & .16 (.02) & .13 (.02) & .15 (.02) & .13 (.02) \\
Science/Arts Male/Female & 0.81* & \cellcolor{gray!20}0.19 & \cellcolor{gray!20}-0.33 & .12 (.02) & .12 (.02) & .12 (.02) & .12 (.02) \\
\hline
 \multicolumn{8}{|c|}{CLIP-ViT-L14-336 EAT Results - With Prompt} \\
 \hline
 Level & \multicolumn{1}{|c|}{Level 1} & \multicolumn{2}{|c|}{Level 2} & \multicolumn{4}{|c|}{Level 3} \\
\hline
EAT (Targets X/Y Attributes A/B) & \multicolumn{1}{|c|}{A,B,X,Y} & \multicolumn{1}{|c|}{A,B,X} & \multicolumn{1}{|c|}{A,B,Y} & \multicolumn{1}{|c|}{A,X} & \multicolumn{1}{|c|}{B,X} & \multicolumn{1}{|c|}{A,Y} & \multicolumn{1}{|c|}{B,Y} \\
 \hline
Flower/Insect P/U25 & 1.84* & \cellcolor{red!20}1.04* & \cellcolor{blue!20}-0.74* & .16 (.02) & .15 (.02) & .15 (.02) & .16 (.02) \\
White/Black P/U25 & \cellcolor{yellow!20}-1.20* & \cellcolor{blue!20}-.94* & \cellcolor{blue!20}-.87* & .17 (.01) & .18 (.01) & .15 (.01) & .16 (.01) \\
Thin/Heavy P/U25 & 1.42* & \cellcolor{gray!20}-0.24 & \cellcolor{blue!20}-0.51* & .16 (.01) & .16 (.01) & .16 (.01) & .16 (.01) \\
Male/Female Career/Family & \cellcolor{yellow!20}0.29 & \cellcolor{red!20}1.25* & \cellcolor{red!20}1.09* & .17 (.02) & .15 (.02) & .16 (.02) & .15 (.02) \\
Science/Arts Male/Female & \cellcolor{yellow!20}0.49 & \cellcolor{gray!20}-0.53 & \cellcolor{gray!20}-0.80 & .13 (.02) & .14 (.02) & .14 (.02) & .15 (.02) \\
\hline
\end{tabular}
\caption{\small ML-EAT results on CLIP-ViT-L14-336 demonstrate that changing the prompt can change the \textit{sign} of the Level 1 effect size, even while EAT patterns (defined by Level 2 effect sizes) remain unchanged. Note that target groups are images and attribute groups are text.} 
\label{tab:CLIP_14_Prompting}
\end{table*}

\subsection{Empirical Analysis: Prompting in CLIP}

In modern zero-shot language-and-image models like CLIP, the choice of prompt can impact the cosine similarities returned by the model. For example, \citet{radford2021learning} suggested adding the prefix ``a photo of a \textbf{class}'' to prompts when using CLIP in the zero-shot setting to improve performance when classifying images. Following the IAT, wherein human reaction times are measured in response to the appearance of pairs of individual words on a screen \cite{greenwald1998measuring}, EATs typically add little context when measuring semantic associations in language technologies, and usually attempt to measure associations between individual words or images. However, some recent research using language-and-image models departs from this \cite{wolfe2023contrastive}, employing longer prompts like that recommended by \citet{radford2021learning} in order to use the model as close to the way it was intended as possible.

Level 3 of the ML-EAT can surface the impact of prompts by revealing underlying CLIP cosine similarities. We study five positively-signed, statistically significant EATs obtained from the CLIP-ViT-L14-336 model using the language stimuli of \citet{caliskan2017semantics} and the image stimuli \citet{steed2021image}, with results in Table \ref{tab:CLIP_14_Prompting}. These EATs include a Flowers-Insects P/U25 test exhibiting the \img{images/ab_div} \textbf{AB-Divergent} EAT pattern (\textit{i.e.,} Flowers associated with Pleasantness, Insects with Unpleasantness); a test of White/Black P/U25 racial bias exhibiting the \img{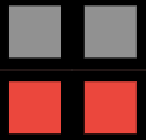} \textbf{B-Uniform} EAT pattern (\textit{i.e.,} both White and Black associated with Unpleasantness); a Thin/Heavy P/U25 test of weight bias exhibiting a \img{images/by_sing} \textbf{BY-Singular} EAT pattern (\textit{i.e.,} Heavy associated with Unpleasantness, Thin with neither Pleasant nor Unpleasant); a Male/Female Career/Family test of gender bias exhibiting an \img{images/a_uni} \textbf{A-Uniform} EAT pattern (\textit{i.e.,} both Male and Female associated with Career); and a Science/Arts Male/Female test of gender bias exhibiting a \img{images/non_dir} \textbf{Non-Directional} EAT pattern (neither target associated with an attribute).

Using prompts with CLIP can affect these measurements. In accordance with the suggestion of \citet{radford2021learning} and in keeping with the the IAT, we used ``a picture that brings to mind [word]'' as the prompt for word stimuli in both the $A$ and $B$ attribute groups (not all image stimuli are photographs, so we prompted with ``picture'' instead of ``photo''). With the prompt, the mean cosine similarity increases for every target-attribute pairing. Observing the impact of prompting helps to understand the variance induced by a particular prompt, and reinforces that these EATs measure \textit{implicit} bias: if the stimuli described exactly what was in the image, the cosine similarities would be higher, as they are proportional to probabilities in CLIP. \looseness=-1

The CLIP measurements make clear another benefit of the ML-EAT. Of the five Level 1 measurements, two are significant only in the absence of the prompt, and the White/Black P/U25 test changes the direction of association. Without the prompt, Level 1 indicates a large, significant valence bias favoring White over Black; with the prompt, Level 1 indicates a large, significant valence bias favoring Black over White. However, the EAT patterns for all five tests, which are based on Level 2 effect sizes, remain the same regardless of whether the prompt is present. This may not always be the case, but it illustrates the importance of being able to observe the measurements that underlie the top level EAT $d$-score. \looseness=-1

\subsection{Empirical Analysis: Anisotropy in GPT-2}\label{sec:gpt2}

Table \ref{GPT2_Full_Results_Table} describes results of the ML-EAT in both the GPT-2 Base model (124 million parameters) and the GPT-2 XL model (1.5 billion parameters) measured using the prompting approach of \citet{may2019measuring}, wherein the model receives the prompt ``This is [stimulus]'' to accord with its training objective. We focus not on the EAT patterns in this case, but on the Level 3 results, which provide evidence of anisotropy (directional uniformity, based on cosine similarity close to 1.0). Given that prior work finds that high anisotropy obscures the semantic properties of contextual word embeddings \cite{mu2018all,timkey2021all}, one might avoid relying on the Level 1 and Level 2 measurements for which these cosine similarities are components. Cosine similarities in GPT-2 XL, on the other hand, may exhibit mild anisotropy, but they also exhibit more variance than the smaller GPT-2 model. While EAT patterns are mostly nondirectional, Level 1 effects are large and significant in the XL model, consistent with results from static word embeddings and with the studies of implicit bias in human subjects.

\begin{table*}[htbp]
\small
\centering
\begin{tabular}
{|l||S[table-format=3.2]|S[table-format=3.2]|S[table-format=3.2]|r|r|r|r|}
 \hline
 \multicolumn{8}{|c|}{GPT-2 Base Full EAT Results} \\
 \hline
 Level & \multicolumn{1}{|c|}{Level 1} & \multicolumn{2}{|c|}{Level 2} & \multicolumn{4}{|c|}{Level 3} \\
\hline
EAT (Targets X/Y Attributes A/B) & \multicolumn{1}{|c|}{A,B,X,Y} & \multicolumn{1}{|c|}{A,B,X} & \multicolumn{1}{|c|}{A,B,Y} & \multicolumn{1}{|c|}{A,X} & \multicolumn{1}{|c|}{B,X} & \multicolumn{1}{|c|}{A,Y} & \multicolumn{1}{|c|}{B,Y} \\
 \hline
Flower/Insect P/U25 & 0.42 & \cellcolor{gray!20}-0.30 & \cellcolor{blue!20}-0.54* & \cellcolor{yellow!20}.99 (.01) & \cellcolor{yellow!20}.99 (.01) & \cellcolor{yellow!20}.99 (.00) & \cellcolor{yellow!20}.99 (.00) \\
Instrument/Weapon P/U25 & -0.20 & \cellcolor{gray!20}-0.16 & \cellcolor{gray!20}-0.01 & \cellcolor{yellow!20}.99 (.00) & \cellcolor{yellow!20}.99 (.00) & \cellcolor{yellow!20}.99 (.01) & \cellcolor{yellow!20}.99 (.01) \\
EA/AA32 P/U25 & 0.23 & \cellcolor{gray!20}0.27 & \cellcolor{gray!20}0.30 & \cellcolor{yellow!20}.97 (.01) & \cellcolor{yellow!20}.97 (.01) & \cellcolor{yellow!20}.98 (.01) & \cellcolor{yellow!20}.98 (.01) \\
EA/AA16 P/U25 & -0.19 & \cellcolor{gray!20}0.18 & \cellcolor{gray!20}0.37 & \cellcolor{yellow!20}.97 (.01) & \cellcolor{yellow!20}.97 (.01) & \cellcolor{yellow!20}.98 (.01) & \cellcolor{yellow!20}.98 (.01) \\
EA/AA16 P/U8 & 0.04 & \cellcolor{gray!20}-0.22 & \cellcolor{gray!20}-0.34 & \cellcolor{yellow!20}.97 (.01) & \cellcolor{yellow!20}.97 (.01) & \cellcolor{yellow!20}.98 (.01) & \cellcolor{yellow!20}.98 (.01) \\
Male/Female Career/Family & 0.08 & \cellcolor{red!20}1.31* & \cellcolor{red!20}1.30* & \cellcolor{yellow!20}.98 (.01) & \cellcolor{yellow!20}.97 (.01) & \cellcolor{yellow!20}.98 (.01) & \cellcolor{yellow!20}.97 (.01) \\
Math/Arts Male/Female & -0.17 & \cellcolor{gray!20}-0.58 & \cellcolor{gray!20}-0.46 & \cellcolor{yellow!20}.99 (.01) & \cellcolor{yellow!20}.99 (.01) & \cellcolor{yellow!20}.99 (.01) & \cellcolor{yellow!20}.99 (.00) \\
Science/Arts Male/Female & -0.44 & \cellcolor{gray!20}-0.24 & \cellcolor{gray!20}-0.18 & \cellcolor{yellow!20}.99 (.01) & \cellcolor{yellow!20}.99 (.01) & \cellcolor{yellow!20}.99 (.01) & \cellcolor{yellow!20}.99 (.00) \\
Mental/Physical Temporary/Permanent & -1.20* & \cellcolor{gray!20}0.81 & \cellcolor{gray!20}0.67 & \cellcolor{yellow!20}.99 (.00) & \cellcolor{yellow!20}.99 (.01) & \cellcolor{yellow!20}.99 (.00) & \cellcolor{yellow!20}.98 (.01) \\
Young/Old P/U8 & -0.44 & \cellcolor{gray!20}-0.31 & \cellcolor{gray!20}-0.20 & \cellcolor{yellow!20}.97 (.01) & \cellcolor{yellow!20}.97 (.01) & \cellcolor{yellow!20}.98 (.01) & \cellcolor{yellow!20}.98 (.01) \\
 \hline
 \multicolumn{8}{|c|}{GPT-2 XL Full EAT Results} \\
 \hline
  Level & \multicolumn{1}{|c|}{Level 1} & \multicolumn{2}{|c|}{Level 2} & \multicolumn{4}{|c|}{Level 3} \\
\hline
EAT (Targets X/Y Attributes A/B) & \multicolumn{1}{|c|}{A,B,X,Y} & \multicolumn{1}{|c|}{A,B,X} & \multicolumn{1}{|c|}{A,B,Y} & \multicolumn{1}{|c|}{A,X} & \multicolumn{1}{|c|}{B,X} & \multicolumn{1}{|c|}{A,Y} & \multicolumn{1}{|c|}{B,Y} \\
\hline
 Flower/Insect P/U25 & 1.65* & \cellcolor{gray!20}0.04 & \cellcolor{blue!20}-0.69* & .29 (.05) & .29 (.06) & .28 (.06) & .31 (.07) \\
Instrument/Weapon P/U25 & 0.95* & \cellcolor{gray!20}-0.02 & \cellcolor{gray!20}-0.28 & .28 (.05) & .28 (.05) & .31 (.06) & .32 (.07) \\
EA/AA32 P/U25 & 0.91* & \cellcolor{gray!20}0.37 & \cellcolor{gray!20}0.24 & .23 (.06) & .21 (.05) & .21 (.06) & .20 (.06) \\
EA/AA16 P/U25 & 0.64* & \cellcolor{gray!20}0.28 & \cellcolor{gray!20}0.18 & .22 (.05) & .21 (.05) & .22 (.07) & .21 (.06) \\
EA/AA16 P/U8 & 0.65* & \cellcolor{gray!20}0.14 & \cellcolor{gray!20}-0.11 & .25 (.04) & .25 (.03) & .24 (.06) & .25 (.06) \\
Male/Female Work/Home & 1.17* & \cellcolor{gray!20}-0.81 & \cellcolor{blue!20}-1.02* & .18 (.04) & .22 (.06) & .18 (.05) & .23 (.06) \\
Math/Arts Male/Female & 0.19 & \cellcolor{gray!20}-0.57 & \cellcolor{gray!20}-0.56 & .31 (.06) & .34 (.06) & .33 (.07) & .36 (.07) \\
Science/Arts Male/Female & 0.33 & \cellcolor{gray!20}-0.36 & \cellcolor{gray!20}-0.40 & .31 (.06) & .33 (.06) & .31 (.07) & .34 (.07) \\
Mental/Physical Temporary/Permanent & 1.52* & \cellcolor{gray!20}0.75 & \cellcolor{gray!20}0.46 & .49 (.09) & .41 (.10) & .34 (.07) & .32 (.05) \\
Young/Old P/U8 & 1.27* & \cellcolor{gray!20}0.56 & \cellcolor{gray!20}-0.00 & .27 (.04) & .25 (.03) & .24 (.05) & .24 (.04) \\
 \hline
\end{tabular}
\caption{\small Level 3 of the ML-EAT surfaces the directional uniformity of contextualized embeddings in GPT-2 base, which results in low variance in cosine similarity and inconsistent Level 1 measurements. However, Level 1 measurements are consistent with societal biases in the XL model, which does not exhibit directional uniformity and has variance in cosine similarities comparable to that observed in static word embeddings. \looseness=-1} 
\label{GPT2_Full_Results_Table}
\end{table*}
\section{Discussion}

Reporting outcomes using the ML-EAT can increase the transparency of research that employs EATs. Rather than attempting to explain the meaning of a single summary statistic and significance test, researchers can drawn on a categorical and visual vocabulary with which to communicate the findings of their work. Given the variance observed in the Level 2 results obtained using stimuli employed in prior studies (even with uniformly large Level 1 effect sizes), describing intermediate results via the ML-EAT might be adopted as a best practice when reporting the outcome of an EAT. 

\subsection{Motivating Interpretable Bias Measurement}

Studies employing the EAT inform how social scientists understand human attitudes \cite{charlesworth2022historical}, and ongoing work uses these measurements to \textit{predict} human bias \cite{bhatia2023predicting}. Ensuring that findings related to societal or domain-specific (\textit{e.g.,} legal, medical, etc.) bias are well-understood is essential not only for the integrity of the scientific record but for the decisions societies may make on the basis of that data. Scholars including \citet{greenwald2022implicit} have suggested that implicit bias might be approached as a public health problem, and mitigated using preventative approaches ``to disable the path from implicit biases to discriminatory outcomes.'' Where an EAT is presented as evidence of an implicit bias in need of redress through public health measures, researchers would be well-served by transparent and interpretable methods. 

Where the ML-EAT informs how intrinsic bias is interpreted, it may also provide information for how bias might be addressed in the embedding itself. In the case of the tests of racial bias (EA/AA Names) in the GloVe embeddings, Level 2 indicates that there is no significant association of African American names with pleasantness or unpleasantness, and Level 3 indicates that this results from a lack of similarity (due to lack of co-occurrence) with words in either attribute $A$ or $B$. This suggests that bias quantified by the EAT might be mitigated via more diverse and representative training data, rather than by aggressively pruning the dataset, a process which can exclude diverse voices \cite{dodge2021documenting}. \looseness=-1

\subsection{Improving the Robustness of the EAT}

The ML-EAT also helps to improve the robustness of EAT-based measurements by introducing a generalization of the SC-EAT in Level 2. Generalizing this test such that the target group can be defined with additional words improves the robustness of single-target tests, which are otherwise dependent on the conceptual representativeness of a single word. Moreover, the ML-EAT is modular enough to use definitions of association other than cosine similarity. For example, one could adopt algebraic definition, following \citet{bolukbasi2016man}, but still use the framework of the ML-EAT to provide transparency at multiple levels of measurement. \looseness=-1

\subsection{Intrinsic Bias and Application Bias}

Prior work documenting limitations of EATs has largely focused on evidence that intrinsic biases do not transfer to downstream tasks \cite{goldfarb2020intrinsic}. While it is not our central concern, we note that, in cases where cosine similarity has an explicitly defined function in a model, biases measured transparently using well-designed EATs will necessarily transfer downstream. This occurs when models like CLIP are used in a zero-shot setting, such that a cosine similarity between text and image is converted into a \textit{probability} for use in classification \cite{radford2021learning}. While intrinsic measurements may not predict application bias in many cases, there remain NLP applications that motivate the transparency and interpretability of EAT measurements.

\subsection{Limitations and Future Work}

While the ML-EAT renders bias more observable, careful curation of stimuli remains necessary to ensure validity \cite{caliskan2022gender}. Ensuring that those stimuli are globally representative is an ongoing challenge for studies employing EATs, as recent research contends that stimuli used in most EATs reflect a western-centric bias that fails to capture biases faced by indigenous populations around the world \cite{yogarajan2023effectiveness}. Moreover, while the ML-EAT is modular with differing mathematical definitions of bias, its levels are not adaptable for some modified versions of the EAT, such as the CEAT, which computes associations using thousands of sentences \cite{guo2021detecting}. Future work might extend the ML-EAT to such tests and uncover new patterns of bias. Finally, our research addresses limitations of observability of bias in EATs, rather than downstream propagation of intrinsic bias. Future work might explore whether ML-EAT measurements can predict application bias. \looseness=-1
\section{Conclusion}

We introduced the ML-EAT, a three-level measurement of intrinsic bias intended to improve the transparency and observability of bias measurement in social science, and appropriate for technologies ranging from static and diachronic word embeddings to zero-shot language-and-image models. We further introduced a taxonomy of nine distinct EAT patterns which we showed occur in prior EATs applied to word embeddings, alongside the EAT-Map, an intuitive visualization for EAT patterns. The ML-EAT provides greater transparency when employing language technologies to understand human minds and societies, an increasing concern as such measurements are used not only to observe human bias, but to predict it \cite{bhatia2023predicting}, and perhaps even to try to prevent it \cite{greenwald2022implicit}. \looseness=-1

\section{Researcher Positionality}

Two of the authors of this research have an extensive background in machine learning and quantitative studies of AI bias, and a third author has extensive experience with human-computer interaction, including statistical studies of deceptive design. We sought to include a variety of perspectives on this project, as the method we intended to develop needed to be both statistically rigorous, yet also approachable and interpretable for researchers hoping to study bias in AI.

\section{Ethical Considerations}

We note that, while we believe reporting results using the ML-EAT will help to make bias research more transparent and interpretable, simply using the ML-EAT rather than the WEAT will not guarantee full transparency in research practices. Researchers must also choose ethical ways of selecting stimuli for WEAT tests, and for determining the number of stimuli to include, which can impact the statistical power of the test. Though pre-registration is sometimes employed for psychological experiments, including those involving word embeddings, the easy availability of language models may render this approach less effective than it is with human subjects experiments, which carry much more notable startup costs. Future work might consider ethical approaches to social scientific experiment design with modern language technologies.

\section{Adverse Impacts}

While we have not produced any new technology in this work, individuals could use our method for ends we have not intended, such as exploiting biases identified with the ML-EAT to further marketing campaigns or to produce misinformation targeted to societal vulnerabilities. We hope and expect that most uses of the method will be to support more transparent and interpretable studies of bias in AI.

\bibliography{custom}
\appendix

\section{Appendix}

\subsection{Use of AI Assistants}

The authors of this research used Github Co-Pilot to improve existing code by adding docstrings and type hints. No assistants were used to assist in the writing of this research paper.

\subsection{EAT Stimuli}\label{sec:stimuli}

The below groups of A, B, X, and Y stimuli are derived from the ten tests of \citet{caliskan2017semantics}, and are employed in the embedding association tests used in this work. Image stimuli are derived from the research of \citet{steed2021image}.

\subsubsection{EAT Text Stimuli}
\noindent \textbf{Pleasant vs. Unpleasant, Flower vs. Insect}

\noindent A (Pleasant): caress, freedom, health, love, peace, cheer, friend, heaven, loyal, pleasure, diamond, gentle, honest, lucky, rainbow, diploma, gift, honor, miracle, sunrise, family, happy, laughter, paradise, vacation

\noindent B (Unpleasant): abuse, crash, filth, murder, sickness, accident, death, grief, poison, stink, assault, disaster, hatred, pollute, tragedy, divorce, jail, poverty, ugly, cancer, kill, rotten, vomit, agony, prison

\noindent X (Flower): aster, clover, hyacinth, marigold, poppy, azalea, crocus, iris, orchid, rose, bluebell, daffodil, lilac, pansy, tulip, buttercup, daisy, lily, peony, violet, carnation, gladiola, magnolia, petunia, zinnia

\noindent Y (Insect): ant, caterpillar, flea, locust, spider, bedbug, centipede, fly, maggot, tarantula, bee, cockroach, gnat, mosquito, termite, beetle, cricket, hornet, moth, wasp, blackfly, dragonfly, horsefly, roach, weevil

\noindent \textbf{Pleasant vs. Unpleasant, Instrument vs. Weapon}

\noindent A (Pleasant): caress, freedom, health, love, peace, cheer, friend, heaven, loyal, pleasure, diamond, gentle, honest, lucky, rainbow, diploma, gift, honor, miracle, sunrise, family, happy, laughter, paradise, vacation

\noindent B (Unpleasant): abuse, crash, filth, murder, sickness, accident, death, grief, poison, stink, assault, disaster, hatred, pollute, tragedy, divorce, jail, poverty, ugly, cancer, kill, rotten, vomit, agony, prison

\noindent X (Instrument): bagpipe, cello, guitar, lute, trombone, banjo, clarinet, harmonica, mandolin, trumpet, bassoon, drum, harp, oboe, tuba, bell, fiddle, harpsichord, piano, viola, bongo, flute, horn, saxophone, violin

\noindent Y (Weapon): arrow, club, gun, missile, spear, axe, dagger, harpoon, pistol, sword, blade, dynamite, hatchet, rifle, tank, bomb, firearm, knife, shotgun, teargas, cannon, grenade, mace, slingshot, whip

\noindent \textbf{Pleasant vs. Unpleasant - Alt, European American vs. African American}

\noindent A (Pleasant): caress, freedom, health, love, peace, cheer, friend, heaven, loyal, pleasure, diamond, gentle, honest, lucky, rainbow, diploma, gift, honor, miracle, sunrise, family, happy, laughter, paradise, vacation

\noindent B (Unpleasant - Alt): abuse, crash, filth, murder, sickness, accident, death, grief, poison, stink, assault, disaster, hatred, pollute, tragedy, divorce, jail, poverty, ugly, cancer, kill, rotten, vomit, bomb, evil

\noindent X (European American): Adam, Harry, Josh, Roger, Alan, Frank, Justin, Ryan, Andrew, Jack, Matthew, Stephen, Brad, Greg, Paul, Jonathan, Peter, Amanda, Courtney, Heather, Melanie, Katie, Betsy, Kristin, Nancy, Stephanie, Ellen, Lauren, Colleen, Emily, Megan, Rachel

\noindent Y (African American): Alonzo, Jamel, Theo, Alphonse, Jerome, Leroy, Torrance, Darnell, Lamar, Lionel, Tyree, Deion, Lamont, Malik, Terrence, Tyrone, Lavon, Marcellus, Wardell, Nichelle, Shereen, Ebony, Latisha, Shaniqua, Jasmine, Tanisha, Tia, Lakisha, Latoya, Yolanda, Malika, Yvette

\noindent \textbf{Pleasant vs. Unpleasant - Alt, European American 2 vs. African American 2}

\noindent A (Pleasant): caress, freedom, health, love, peace, cheer, friend, heaven, loyal, pleasure, diamond, gentle, honest, lucky, rainbow, diploma, gift, honor, miracle, sunrise, family, happy, laughter, paradise, vacation

\noindent B (Unpleasant - Alt): abuse, crash, filth, murder, sickness, accident, death, grief, poison, stink, assault, disaster, hatred, pollute, tragedy, divorce, jail, poverty, ugly, cancer, kill, rotten, vomit, bomb, evil

\noindent X (European American 2): Brad, Brendan, Geoffrey, Greg, Brett, Matthew, Neil, Todd, Allison, Anne, Carrie, Emily, Jill, Laurie, Meredith, Sarah

\noindent Y (African American 2): Darnell, Hakim, Jermaine, Kareem, Jamal, Leroy, Rasheed, Tyrone, Aisha, Ebony, Keisha, Kenya, Lakisha, Latoya, Tamika, Tanisha

\noindent \textbf{Pleasant 2 vs. Unpleasant 2, European American 2 vs. African American 2}

\noindent A (Pleasant 2): joy, love, peace, wonderful, pleasure, friend, laughter, happy

\noindent B (Unpleasant 2): agony, terrible, horrible, nasty, evil, war, awful, failure

\noindent X (European American 2): Brad, Brendan, Geoffrey, Greg, Brett, Matthew, Neil, Todd, Allison, Anne, Carrie, Emily, Jill, Laurie, Meredith, Sarah

\noindent Y (African American 2): Darnell, Hakim, Jermaine, Kareem, Jamal, Leroy, Rasheed, Tyrone, Aisha, Ebony, Keisha, Kenya, Lakisha, Latoya, Tamika, Tanisha

\noindent \textbf{Career vs. Domestic, Male Name vs. Female Name}

\noindent A (Career): executive, management, professional, corporation, salary, office, business, career

\noindent B (Domestic): home, parents, children, family, cousins, marriage, wedding, relatives

\noindent X (Male Name): John, Paul, Mike, Kevin, Steve, Greg, Jeff, Bill

\noindent Y (Female Name): Amy, Joan, Lisa, Sarah, Diana, Kate, Ann, Donna

\noindent \textbf{Male Terms vs. Female Terms, Math vs. Art}

\noindent A (Male Terms): male, man, boy, brother, he, him, his, son

\noindent B (Female Terms): female, woman, girl, sister, she, her, hers, daughter

\noindent X (Math): math, algebra, geometry, calculus, equations, computation, numbers, addition

\noindent Y (Art): poetry, art, dance, literature, novel, symphony, drama, sculpture

\noindent \textbf{Male Terms 2 vs. Female Terms 2, Science vs. Art 2}

\noindent A (Male Terms 2): brother, father, uncle, grandfather, son, he, his, him

\noindent B (Female Terms 2): sister, mother, aunt, grandmother, daughter, she, hers, her

\noindent X (Science): science, technology, physics, chemistry, Einstein, NASA, experiment, astronomy

\noindent Y (Art 2): poetry, art, Shakespeare, dance, literature, novel, symphony, drama

\noindent \textbf{Temporary vs. Permanent, Mental vs. Physical}

\noindent A (Temporary): impermanent, unstable, variable, fleeting, short-term, brief, occasional

\noindent B (Permanent): stable, always, constant, persistent, chronic, prolonged, forever

\noindent X (Mental): sad, hopeless, gloomy, tearful, miserable, depressed

\noindent Y (Physical): sick, illness, influenza, disease, virus, cancer

\noindent \textbf{Pleasant 2 vs. Unpleasant 2, Young vs. Old}

\noindent A (Pleasant 2): joy, love, peace, wonderful, pleasure, friend, laughter, happy

\noindent B (Unpleasant 2): agony, terrible, horrible, nasty, evil, war, awful, failure

\noindent X (Young): Tiffany, Michelle, Cindy, Kristy, Brad, Eric, Joey, Billy

\noindent Y (Old): Ethel, Bernice, Gertrude, Agnes, Cecil, Wilbert, Mortimer, Edgar

\subsection{Comprehensive ML-EAT Results}

The tables below contain full ML-EAT results from the Google News embeddings, on which \citet{caliskan2017semantics} evaluated in their supplementary materials, along with GPT-2 ML-EAT tests, inserting the EAT stimuli in Appendix \ref{sec:stimuli} into sentence prompts in accordance with the methodology described in Section \ref{sec:gpt2}. Note that all embeddings are obtained from the top layer of the GPT-2 model in question. The tables below the GPT-2 results include CLIP ML-EAT measurements, using the Pleasant/Unpleasant EAT stimuli from Appendix \ref{sec:stimuli} as attributes, and adopting the image stimuli of \citet{steed2021image} as target groups.

\begin{table*}[htbp]
\centering
\small
\begin{tabular}
{|l||S[table-format=3.2]|S[table-format=3.2]|S[table-format=3.2]|r|r|r|r|}
 \hline
 \multicolumn{8}{|c|}{Multilevel Embedding Association Test: Google News Embeddings} \\
 \hline
 Level & \multicolumn{1}{|c|}{Level 1} & \multicolumn{2}{|c|}{Level 2} & \multicolumn{4}{|c|}{Level 3} \\
\hline
EAT (Targets X/Y Attributes A/B) & \multicolumn{1}{|c|}{A,B,X,Y} & \multicolumn{1}{|c|}{A,B,X} & \multicolumn{1}{|c|}{A,B,Y} & \multicolumn{1}{|c|}{A,X} & \multicolumn{1}{|c|}{B,X} & \multicolumn{1}{|c|}{A,Y} & \multicolumn{1}{|c|}{B,Y} \\
 \hline
Flower/Insect P/U25 & 1.54* & \cellcolor{red!20}0.78* & \cellcolor{gray!20}-0.28 & 0.11 (0.08) & 0.07 (0.05) & 0.08 (0.07) & 0.09 (0.07) \\
Instrument/Weapon P/U25 & 1.63* & \cellcolor{red!20}0.96* & \cellcolor{gray!20}-0.42 & 0.10 (0.07) & 0.05 (0.05) & 0.07 (0.06) & 0.09 (0.08) \\
EA/AA32 P/U25 & 0.58* & \cellcolor{gray!20}0.32 & \cellcolor{gray!20}-0.03 & 0.06 (0.05) & 0.05 (0.04) & 0.06 (0.05) & 0.06 (0.05) \\
EA/AA16 P/U25 & 1.24* & \cellcolor{gray!20}0.43 & \cellcolor{gray!20}-0.18 & 0.06 (0.05) & 0.05 (0.04) & 0.06 (0.05) & 0.07 (0.06) \\
EA/AA18 P/U8 & 0.72* & \cellcolor{gray!20}0.61 & \cellcolor{gray!20}0.35 & 0.09 (0.06) & 0.06 (0.05) & 0.08 (0.06) & 0.07 (0.04) \\
Male/Female Career/Family & 1.89* & \cellcolor{red!20}1.52* & \cellcolor{blue!20}-1.37* & 0.11 (0.05) & 0.01 (0.04) & 0.07 (0.05) & 0.14 (0.05) \\
Math/Arts Male/Female & 0.97* & \cellcolor{gray!20}-0.48 & \cellcolor{blue!20}-1.22* & 0.03 (0.05) & 0.04 (0.06) & 0.08 (0.05) & 0.12 (0.06) \\
Science/Arts Male/Female & 1.24* & \cellcolor{gray!20}-0.09 & \cellcolor{blue!20}-1.36* & 0.07 (0.05) & 0.07 (0.05) & 0.07 (0.04) & 0.12 (0.06) \\
Mental/Physical Temp/Perm & 1.30* & \cellcolor{gray!20}-0.09 & \cellcolor{blue!20}-1.04* & 0.16 (0.10) & 0.16 (0.09) & 0.06 (0.06) & 0.12 (0.09) \\
Young/Old P/U8 & -0.09 & \cellcolor{gray!20}0.56 & \cellcolor{red!20}1.07* & 0.10 (0.07) & 0.07 (0.06) & 0.10 (0.06) & 0.07 (0.03) \\\hline
\end{tabular}
\caption{\small Full results of the ML-EAT on the Google News embeddings, which were assessed by \citet{caliskan2017semantics} in their supplementary materials.} 
\label{MLEAT_Results_Table_2}
\end{table*}

\begin{table*}[htbp]
\centering
\small
\begin{tabular}
{|l||S[table-format=3.2]|S[table-format=3.2]|S[table-format=3.2]|r|r|r|r|}
 \hline
 \multicolumn{8}{|c|}{GPT-2 Base Full EAT Results} \\
 \hline
 Level & \multicolumn{1}{|c|}{Level 1} & \multicolumn{2}{|c|}{Level 2} & \multicolumn{4}{|c|}{Level 3} \\
\hline
EAT (Targets X/Y Attributes A/B) & \multicolumn{1}{|c|}{A,B,X,Y} & \multicolumn{1}{|c|}{A,B,X} & \multicolumn{1}{|c|}{A,B,Y} & \multicolumn{1}{|c|}{A,X} & \multicolumn{1}{|c|}{B,X} & \multicolumn{1}{|c|}{A,Y} & \multicolumn{1}{|c|}{B,Y} \\
 \hline
Flower/Insect P/U25 & 0.42 & \cellcolor{gray!20}-0.30 & \cellcolor{blue!20}-0.54* & .99 (.01) & .99 (.01) & .99 (.00) & .99 (.00) \\
Instrument/Weapon P/U25 & -0.20 & \cellcolor{gray!20}-0.16 & \cellcolor{gray!20}-0.01 & .99 (.00) & .99 (.00) & .99 (.01) & .99 (.01) \\
EA/AA32 P/U25 & 0.23 & \cellcolor{gray!20}0.27 & \cellcolor{gray!20}0.30 & .97 (.01) & .97 (.01) & .98 (.01) & .98 (.01) \\
EA/AA16 P/U25 & -0.19 & \cellcolor{gray!20}0.18 & \cellcolor{gray!20}0.37 & .97 (.01) & .97 (.01) & .98 (.01) & .98 (.01) \\
EA/AA16 P/U8 & 0.04 & \cellcolor{gray!20}-0.22 & \cellcolor{gray!20}-0.34 & .97 (.01) & .97 (.01) & .98 (.01) & .98 (.01) \\
Male/Female Work/Home & 0.08 & \cellcolor{red!20}1.31* & \cellcolor{red!20}1.30* & .98 (.01) & .97 (.01) & .98 (.01) & .97 (.01) \\
Math/Arts Male/Female & -0.17 & \cellcolor{gray!20}-0.58 & \cellcolor{gray!20}-0.46 & .99 (.01) & .99 (.01) & .99 (.01) & .99 (.00) \\
Science/Arts Male/Female & -0.44 & \cellcolor{gray!20}-0.24 & \cellcolor{gray!20}-0.18 & .99 (.01) & .99 (.01) & .99 (.01) & .99 (.00) \\
Mental/Physical Temporary/Permanent & -1.20* & \cellcolor{gray!20}0.81 & \cellcolor{gray!20}0.67 & .99 (.00) & .99 (.01) & .99 (.00) & .98 (.01) \\
Young/Old P/U8 & -0.44 & \cellcolor{gray!20}-0.31 & \cellcolor{gray!20}-0.20 & .97 (.01) & .97 (.01) & .98 (.01) & .98 (.01) \\
 \hline
\end{tabular}
\caption{\small Full results on GPT-2 Base.} 
\label{GPT2_Base_Full_Results_Table}
\end{table*}

\begin{table*}[htbp]
\small
\centering
\begin{tabular}
{|l||S[table-format=3.2]|S[table-format=3.2]|S[table-format=3.2]|r|r|r|r|}
 \hline
 \multicolumn{8}{|c|}{GPT-2 Medium Full EAT Results} \\
 \hline
 Level & \multicolumn{1}{|c|}{Level 1} & \multicolumn{2}{|c|}{Level 2} & \multicolumn{4}{|c|}{Level 3} \\
\hline
EAT (Targets X/Y Attributes A/B) & \multicolumn{1}{|c|}{A,B,X,Y} & \multicolumn{1}{|c|}{A,B,X} & \multicolumn{1}{|c|}{A,B,Y} & \multicolumn{1}{|c|}{A,X} & \multicolumn{1}{|c|}{B,X} & \multicolumn{1}{|c|}{A,Y} & \multicolumn{1}{|c|}{B,Y} \\
 \hline
Flower/Insect P/U25 & 0.57* & \cellcolor{gray!20}0.00 & \cellcolor{gray!20}-0.05 & .99 (.01) & .99 (.01) & .98 (.02) & .98 (.02) \\
Instrument/Weapon P/U25 & 0.36 & \cellcolor{gray!20}0.01 & \cellcolor{gray!20}-0.01 & .99 (.01) & .99 (.01) & .99 (.01) & .99 (.01) \\
EA/AA32 P/U25 & -0.68* & \cellcolor{gray!20}0.10 & \cellcolor{gray!20}0.17 & .99 (.00) & .99 (.00) & .99 (.01) & .99 (.01) \\
EA/AA16 P/U25 & -0.66* & \cellcolor{gray!20}0.11 & \cellcolor{gray!20}0.16 & .99 (.01) & .99 (.01) & .99 (.01) & .99 (.01) \\
EA/AA16 P/U8 & 1.08* & \cellcolor{gray!20}0.41 & \cellcolor{gray!20}-0.41 & .99 (.01) & .99 (.01) & .99 (.01) & .99 (.00) \\
Male/Female Work/Home & -0.03 & \cellcolor{gray!20}0.77 & \cellcolor{gray!20}0.81 & .99 (.00) & .98 (.01) & .99 (.01) & .98 (.01) \\
Math/Arts Male/Female & 0.50 & \cellcolor{gray!20}-0.36 & \cellcolor{gray!20}-0.39 & .99 (.01) & .99 (.01) & .99 (.01) & .99 (.01) \\
Science/Arts Male/Female & 0.11 & \cellcolor{gray!20}-0.32 & \cellcolor{gray!20}-0.33 & .99 (.01) & .99 (.01) & .99 (.01) & .99 (.01) \\
Mental/Physical Temporary/Permanent & -1.40* & \cellcolor{red!20}0.89* & \cellcolor{gray!20}0.56 & .00 (.00) & .99 (.00) & .99 (.00) & .99 (.01) \\
Young/Old P/U8 & -0.31 & \cellcolor{gray!20}-0.12 & \cellcolor{gray!20}0.24 & .99 (.00) & .99 (.00) & .98 (.01) & .98 (.01) \\
\hline
\end{tabular}
\caption{\small Full results on GPT-2 Medium.} 
\label{GPT2_Medium_Full_Results_Table}
\end{table*}

\begin{table*}[htbp]
\centering
\small
\begin{tabular}
{|l||S[table-format=3.2]|S[table-format=3.2]|S[table-format=3.2]|r|r|r|r|}
 \hline
 \multicolumn{8}{|c|}{GPT-2 Large Full EAT Results} \\
 \hline
 Level & \multicolumn{1}{|c|}{Level 1} & \multicolumn{2}{|c|}{Level 2} & \multicolumn{4}{|c|}{Level 3} \\
\hline
EAT (Targets X/Y Attributes A/B) & \multicolumn{1}{|c|}{A,B,X,Y} & \multicolumn{1}{|c|}{A,B,X} & \multicolumn{1}{|c|}{A,B,Y} & \multicolumn{1}{|c|}{A,X} & \multicolumn{1}{|c|}{B,X} & \multicolumn{1}{|c|}{A,Y} & \multicolumn{1}{|c|}{B,Y} \\
 \hline
Flower/Insect P/U25 & 1.67* & \cellcolor{gray!20}0.10 & \cellcolor{blue!20}-0.71* & .32 (.06) & .32 (.07) & .30 (.06) & .34 (.07) \\
Instrument/Weapon P/U25 & 1.40* & \cellcolor{gray!20}0.02 & \cellcolor{gray!20}-0.45 & .30 (.05) & .30 (.05) & .32 (.06) & .34 (.07) \\
EA/AA32 P/U25 & 0.15 & \cellcolor{gray!20}0.16 & \cellcolor{gray!20}0.14 & .25 (.06) & .24 (.05) & .25 (.07) & .25 (.07) \\
EA/AA16 P/U25 & 0.05 & \cellcolor{gray!20}0.13 & \cellcolor{gray!20}0.12 & .24 (.05) & .24 (.05) & .26 (.07) & .25 (.07) \\
EA/AA16 P/U8 & 0.23 & \cellcolor{gray!20}-0.01 & \cellcolor{gray!20}-0.09 & .28 (.04) & .28 (.03) & .29 (.06) & .29 (.06) \\
Male/Female Work/Home & 1.22* & \cellcolor{blue!20}-1.00* & \cellcolor{blue!20}-1.26* & .19 (.05) & .25 (.05) & .20 (.06) & .28 (.07) \\
Math/Arts Male/Female & 0.18 & \cellcolor{gray!20}-0.71 & \cellcolor{blue!20}-0.86* & .34 (.06) & .38 (.07) & .34 (.06) & .38 (.06) \\
Science/Arts Male/Female & 0.70 & \cellcolor{gray!20}-0.37 & \cellcolor{gray!20}-0.52 & .34 (.06) & .36 (.06) & .33 (.06) & .36 (.06) \\
Mental/Physical Temporary/Permanent & 1.37* & \cellcolor{gray!20}0.83 & \cellcolor{gray!20}0.84 & .51 (.09) & .43 (.08) & .38 (.06) & .34 (.06) \\
Young/Old P/U8 & -0.19 & \cellcolor{gray!20}0.10 & \cellcolor{gray!20}0.19 & .29 (.05) & .29 (.04) & .26 (.04) & .25 (.03) \\
\hline
\end{tabular}
\caption{\small Full results on GPT-2 Large.} 
\label{GPT2_Large_Full_Results_Table}
\end{table*}

\begin{table*}[htbp]
\centering
\small
\begin{tabular}
{|l||S[table-format=3.2]|S[table-format=3.2]|S[table-format=3.2]|r|r|r|r|}
 \hline
 \multicolumn{8}{|c|}{GPT-2 XL Full EAT Results} \\
 \hline
 Level & \multicolumn{1}{|c|}{Level 1} & \multicolumn{2}{|c|}{Level 2} & \multicolumn{4}{|c|}{Level 3} \\
\hline
EAT (Targets X/Y Attributes A/B) & \multicolumn{1}{|c|}{A,B,X,Y} & \multicolumn{1}{|c|}{A,B,X} & \multicolumn{1}{|c|}{A,B,Y} & \multicolumn{1}{|c|}{A,X} & \multicolumn{1}{|c|}{B,X} & \multicolumn{1}{|c|}{A,Y} & \multicolumn{1}{|c|}{B,Y} \\
 \hline
Flower/Insect P/U25 & 1.65* & \cellcolor{gray!20}0.04 & \cellcolor{blue!20}-0.69* & .29 (.05) & .29 (.06) & .28 (.06) & .31 (.07) \\
Instrument/Weapon P/U25 & 0.95* & \cellcolor{gray!20}-0.02 & \cellcolor{gray!20}-0.28 & .28 (.05) & .28 (.05) & .31 (.06) & .32 (.07) \\
EA/AA32 P/U25 & 0.91* & \cellcolor{gray!20}0.37 & \cellcolor{gray!20}0.24 & .23 (.06) & .21 (.05) & .21 (.06) & .20 (.06) \\
EA/AA16 P/U25 & 0.64* & \cellcolor{gray!20}0.28 &\cellcolor{gray!20} 0.18 & .22 (.05) & .21 (.05) & .22 (.07) & .21 (.06) \\
EA/AA16 P/U8 & 0.65* & \cellcolor{gray!20}0.14 & \cellcolor{gray!20}-0.11 & .25 (.04) & .25 (.03) & .24 (.06) & .25 (.06) \\
Male/Female Work/Home & 1.17* & \cellcolor{gray!20}-0.81 & \cellcolor{blue!20}-1.02* & .18 (.04) & .22 (.06) & .18 (.05) & .23 (.06) \\
Math/Arts Male/Female & 0.19 & \cellcolor{gray!20}-0.57 & \cellcolor{gray!20}-0.56 & .31 (.06) & .34 (.06) & .33 (.07) & .36 (.07) \\
Science/Arts Male/Female & 0.33 & \cellcolor{gray!20}-0.36 & \cellcolor{gray!20}-0.40 & .31 (.06) & .33 (.06) & .31 (.07) & .34 (.07) \\
Mental/Physical Temporary/Permanent & 1.52* & \cellcolor{gray!20}0.75 & \cellcolor{gray!20}0.46 & .49 (.09) & .41 (.10) & .34 (.07) & .32 (.05) \\
Young/Old P/U8 & 1.27* & \cellcolor{gray!20}0.56 & \cellcolor{gray!20}0.00 & .27 (.04) & .25 (.03) & .24 (.05) & .24 (.04) \\
\hline
\end{tabular}
\caption{\small Full results on GPT-2 XL.} 
\label{GPT2_XL_Full_Results_Table}
\end{table*}

\begin{table*}[htbp]
\centering
\small
\begin{tabular}
{|l||S[table-format=3.2]|S[table-format=3.2]|S[table-format=3.2]|r|r|r|r|}
 \hline
 \multicolumn{8}{|c|}{CLIP-ViT-B32 Full EAT Results} \\
 \hline
 Level & \multicolumn{1}{|c|}{Level 1} & \multicolumn{2}{|c|}{Level 2} & \multicolumn{4}{|c|}{Level 3} \\
\hline
EAT (Targets X/Y Attributes A/B) & \multicolumn{1}{|c|}{A,B,X,Y} & \multicolumn{1}{|c|}{A,B,X} & \multicolumn{1}{|c|}{A,B,Y} & \multicolumn{1}{|c|}{A,X} & \multicolumn{1}{|c|}{B,X} & \multicolumn{1}{|c|}{A,Y} & \multicolumn{1}{|c|}{B,Y} \\
 \hline
Flower/Insect P/U25 & 1.89* & \cellcolor{red!20}0.99* & \cellcolor{blue!20}-0.61* & .20 (.02) & .19 (.01) & .19 (.02) & .20 (.02) \\
Young/Old P/U25 & 1.37* & \cellcolor{blue!20}-1.06* & \cellcolor{blue!20}-1.26* & .20 (.02) & .22 (.01) & .20 (.02) & .22 (.01) \\
Non-Arab-Muslim/Arab-Muslim P/U25 & 0.16 & \cellcolor{gray!20}0.26 & \cellcolor{gray!20}0.09 & .20 (.02) & .19 (.01) & .19 (.02) & .19 (.02) \\
Abled/Disabled P/U25 & 0.13 & \cellcolor{blue!20}-0.91* & \cellcolor{blue!20}-1.05* & .20 (.01) & .21 (.01) & .21 (.01) & .22 (.01) \\
White/Black P/U25 & -1.01* & \cellcolor{blue!20}-1.15* & \cellcolor{blue!20}-0.97* & .21 (.02) & .22 (.01) & .20 (.01) & .21 (.01) \\
Christianity/Judaism P/U25 & -0.25 & \cellcolor{gray!20}0.12 & \cellcolor{gray!20}0.20 & .19 (.02) & .19 (.02) & .19 (.02) & .19 (.01) \\
Straight/Gay P/U25 & 0.46 & \cellcolor{red!20}0.50* & \cellcolor{gray!20}0.25 & .19 (.02) & .18 (.02) & .19 (.02) & .19 (.02) \\
Light-Skin/Dark-Skin P/U25 & -0.63 & \cellcolor{blue!20}-1.02* & \cellcolor{blue!20}-0.92* & .17 (.02) & .19 (.01) & .18 (.02) & .19 (.01) \\
Thin/Heavy P/U25 & 1.31* & \cellcolor{gray!20}-0.39 & \cellcolor{blue!20}-0.70* & .20 (.01) & .20 (.01) & .20 (.01) & .21 (.01) \\
Male/Female Career/Family & 0.50* & \cellcolor{red!20}1.59* & \cellcolor{red!20}1.39* & .21 (.02) & .19 (.02) & .20 (.02) & .18 (.02) \\
Science/Arts Male/Female & 1.09* & \cellcolor{gray!20}0.66 & \cellcolor{gray!20}0.35 & .18 (.02) & .17 (.02) & .18 (.02) & .17 (.02) \\\hline
\end{tabular}
\caption{\small Full results on CLIP-ViT-B32.} 
\label{CLIP_B32_Full_Results_Table}
\end{table*}

\begin{table*}[htbp]
\centering
\small
\begin{tabular}
{|l||S[table-format=3.2]|S[table-format=3.2]|S[table-format=3.2]|r|r|r|r|}
 \hline
 \multicolumn{8}{|c|}{CLIP-ViT-B16 Full EAT Results} \\
 \hline
 Level & \multicolumn{1}{|c|}{Level 1} & \multicolumn{2}{|c|}{Level 2} & \multicolumn{4}{|c|}{Level 3} \\
\hline
EAT (Targets X/Y Attributes A/B) & \multicolumn{1}{|c|}{A,B,X,Y} & \multicolumn{1}{|c|}{A,B,X} & \multicolumn{1}{|c|}{A,B,Y} & \multicolumn{1}{|c|}{A,X} & \multicolumn{1}{|c|}{B,X} & \multicolumn{1}{|c|}{A,Y} & \multicolumn{1}{|c|}{B,Y} \\
 \hline
Flower/Insect P/U25 & 1.84* & \cellcolor{red!20}0.64* & \cellcolor{blue!20}-1.08* & .20 (.02) & .19 (.01) & .18 (.02) & .20 (.02) \\
Young/Old P/U25 & -0.70 & \cellcolor{blue!20}-1.25* & \cellcolor{blue!20}-1.13* & .20 (.01) & .22 (.01) & .21 (.01) & .22 (.01) \\
Non-Arab-Muslim/Arab-Muslim P/U25 & -0.19 & \cellcolor{gray!20}-0.14 & \cellcolor{gray!20}0.02 & .19 (.02) & .20 (.01) & .19 (.02) & .19 (.02) \\
Abled/Disabled P/U25 & 0.19 & \cellcolor{blue!20}-0.53* & \cellcolor{blue!20}-0.69* & .19 (.02) & .20 (.02) & .21 (.02) & .22 (.01) \\
White/Black P/U25 & 0.25 & \cellcolor{blue!20}-1.19* & \cellcolor{blue!20}-1.26* & .20 (.01) & .22 (.01) & .20 (.02) & .22 (.01) \\
Christianity/Judaism P/U25 & -0.12 & \cellcolor{gray!20}-0.18 & \cellcolor{gray!20}-0.10 & .18 (.02) & .19 (.02) & .19 (.02) & .19 (.02) \\
Straight/Gay P/U25 & 0.10 & \cellcolor{gray!20}0.46 & \cellcolor{gray!20}0.46 & .19 (.02) & .19 (.01) & .19 (.02) & .19 (.02) \\
Light-Skin/Dark-Skin P/U25 & -0.08 & \cellcolor{blue!20}-0.98* & \cellcolor{blue!20}-1.06* & .18 (.02) & .20 (.01) & .18 (.02) & .20 (.01) \\
Thin/Heavy P/U25 & 1.42* & \cellcolor{gray!20}-0.05 & \cellcolor{blue!20}-0.50* & .20 (.01) & .20 (.01) & .20 (.01) & .21 (.01) \\
Male/Female Career/Family & 0.36 & \cellcolor{red!20}1.61* & \cellcolor{red!20}1.51* & .21 (.02) & .18 (.02) & .21 (.02) & .18 (.02) \\
Science/Arts Male/Female & 0.76* & \cellcolor{gray!20}0.38 & \cellcolor{gray!20}0.18 & .18 (.02) & .17 (.02) & .18 (.02) & .18 (.01) \\
\hline
\end{tabular}
\caption{\small Full results on CLIP-ViT-B16.} 
\label{CLIP_B16_Full_Results_Table}
\end{table*}

\begin{table*}[htbp]
\centering
\small
\begin{tabular}
{|l||S[table-format=3.2]|S[table-format=3.2]|S[table-format=3.2]|r|r|r|r|}
 \hline
 \multicolumn{8}{|c|}{CLIP-ViT-L14 Full EAT Results} \\
 \hline
 Level & \multicolumn{1}{|c|}{Level 1} & \multicolumn{2}{|c|}{Level 2} & \multicolumn{4}{|c|}{Level 3} \\
\hline
EAT (Targets X/Y Attributes A/B) & \multicolumn{1}{|c|}{A,B,X,Y} & \multicolumn{1}{|c|}{A,B,X} & \multicolumn{1}{|c|}{A,B,Y} & \multicolumn{1}{|c|}{A,X} & \multicolumn{1}{|c|}{B,X} & \multicolumn{1}{|c|}{A,Y} & \multicolumn{1}{|c|}{B,Y} \\
 \hline
Flower/Insect P/U25 & 1.88* & \cellcolor{red!20}0.76* & \cellcolor{blue!20}-1.03* & .14 (.02) & .13 (.01) & .12 (.02) & .14 (.02) \\
Young/Old P/U25 & -1.29* & \cellcolor{blue!20}-1.29* & \cellcolor{blue!20}-1.02* & .14 (.01) & .16 (.02) & .15 (.01) & .16 (.01) \\
Non-Arab-Muslim/Arab-Muslim P/U25 & 0.39 & \cellcolor{gray!20}0.23 & \cellcolor{gray!20}-0.05 & .14 (.02) & .13 (.02) & .13 (.02) & .13 (.02) \\
Abled/Disabled P/U25 & -0.76 & \cellcolor{blue!20}-0.72* & \cellcolor{blue!20}-0.58* & .14 (.02) & .15 (.02) & .15 (.02) & .15 (.02) \\
White/Black P/U25 & 0.52 & \cellcolor{blue!20}-1.25* & \cellcolor{blue!20}-1.31* & .16 (.01) & .18 (.01) & .14 (.02) & .16 (.02) \\
Christianity/Judaism P/U25 & 0.33 & \cellcolor{gray!20}-0.22 & \cellcolor{gray!20}-0.39 & .12 (.02) & .13 (.02) & .13 (.02) & .13 (.02) \\
Straight/Gay P/U25 & 0.07 & \cellcolor{gray!20}0.28 & \cellcolor{gray!20}0.23 & .13 (.02) & .13 (.02) & .13 (.02) & .13 (.02) \\
Light-Skin/Dark-Skin P/U25 & 0.30 & \cellcolor{blue!20}-0.77* & \cellcolor{blue!20}-0.80* & .13 (.02) & .14 (.02) & .13 (.02) & .15 (.02) \\
Thin/Heavy P/U25 & 1.46* & \cellcolor{gray!20}-0.42 & \cellcolor{blue!20}-0.89* & .15 (.01) & .15 (.01) & .13 (.02) & .15 (.01) \\
Male/Female Career/Family & 0.43* & \cellcolor{red!20}1.65* & \cellcolor{red!20}1.54* & .15 (.02) & .12 (.02) & .15 (.02) & .12 (.02) \\
Science/Arts Male/Female & 0.85* & \cellcolor{gray!20}0.21 & \cellcolor{gray!20}-0.27 & .11 (.03) & .11 (.03) & .12 (.02) & .12 (.02) \\
\hline
\end{tabular}
\caption{\small Full results on CLIP-ViT-L14.} 
\label{CLIP_L14_Full_Results_Table}
\end{table*}

\begin{table*}[htbp]
\centering
\small
\begin{tabular}
{|l||S[table-format=3.2]|S[table-format=3.2]|S[table-format=3.2]|r|r|r|r|}
 \hline
 \multicolumn{8}{|c|}{CLIP-ViT-L14-336 Full EAT Results} \\
 \hline
 Level & \multicolumn{1}{|c|}{Level 1} & \multicolumn{2}{|c|}{Level 2} & \multicolumn{4}{|c|}{Level 3} \\
\hline
EAT (Targets X/Y Attributes A/B) & \multicolumn{1}{|c|}{A,B,X,Y} & \multicolumn{1}{|c|}{A,B,X} & \multicolumn{1}{|c|}{A,B,Y} & \multicolumn{1}{|c|}{A,X} & \multicolumn{1}{|c|}{B,X} & \multicolumn{1}{|c|}{A,Y} & \multicolumn{1}{|c|}{B,Y} \\
 \hline
Flower/Insect P/U25 & 1.88* & \cellcolor{red!20}0.87* & \cellcolor{blue!20}-1.03* & .15 (.02) & .14 (.02) & .13 (.02) & .15 (.02) \\
Young/Old P/U25 & -1.43* & \cellcolor{blue!20}-1.38* & \cellcolor{blue!20}-0.98* & .14 (.01) & .16 (.02) & .15 (.01) & .17 (.01) \\
Non-Arab-Muslim/Arab-Muslim P/U25 & 0.46 & \cellcolor{gray!20}0.41 & \cellcolor{gray!20}0.02 & .14 (.02) & .14 (.02) & .13 (.02) & .13 (.02) \\
Abled/Disabled P/U25 & -0.79 & \cellcolor{blue!20}-0.78* & \cellcolor{blue!20}-0.65* & .14 (.02) & .15 (.02) & .15 (.02) & .16 (.02) \\
White/Black P/U25 & 1.05* & \cellcolor{blue!20}-1.27* & \cellcolor{blue!20}-1.40* & .16 (.01) & .18 (.01) & .14 (.02) & .16 (.02) \\
Christianity/Judaism P/U25 & 0.24 & \cellcolor{gray!20}-0.15 & \cellcolor{gray!20}-0.30 & .13 (.02) & .13 (.02) & .13 (.02) & .13 (.02) \\
Straight/Gay P/U25 & -0.28 & \cellcolor{gray!20}0.21 & \cellcolor{gray!20}0.39 & .13 (.02) & .13 (.02) & .14 (.02) & .13 (.02) \\
Light-Skin/Dark-Skin P/U25 & 0.43 & \cellcolor{blue!20}-0.92* & \cellcolor{blue!20}-0.92* & .13 (.02) & .15 (.02) & .13 (.02) & .15 (.02) \\
Thin/Heavy P/U25 & 1.58* & \cellcolor{gray!20}-0.44 & \cellcolor{blue!20}-0.90* & .15 (.01) & .16 (.01) & .14 (.02) & .15 (.01) \\
Male/Female Career/Family & 0.46* & \cellcolor{red!20}1.66* & \cellcolor{red!20}1.54* & .16 (.02) & .13 (.02) & .15 (.02) & .13 (.02) \\
Science/Arts Male/Female & 0.81* & \cellcolor{gray!20}0.19 & \cellcolor{gray!20}-0.33 & .12 (.02) & .12 (.02) & .12 (.02) & .12 (.02) \\
\hline
\end{tabular}
\caption{Full results on CLIP-ViT-L14-336.} 
\label{CLIP_14_Full_Results_Table}
\end{table*}

\end{document}